\documentclass[lettersize,journal]{IEEEtran}
\usepackage{amsmath,amsfonts}
\usepackage{algorithmic}
\usepackage{array}
\usepackage[caption=false,font=normalsize,labelfont=sf,textfont=sf]{subfig}
\usepackage{textcomp}
\usepackage{stfloats}
\usepackage{url}
\usepackage{verbatim}
\usepackage{graphicx}
\usepackage{cite}
\usepackage{xcolor}
\usepackage{booktabs}
\usepackage[lined,boxed,commentsnumbered,ruled]{algorithm2e}
\usepackage{multirow}
\usepackage{tcolorbox}

\tcbset{
    colframe=blue!50!black,
    colback=yellow!20,
    boxrule=0.5mm,
    arc=1mm,
    on line,
    size=fbox
}

\newtheorem{definition}{Definition}

\begin{document}

\title{It's Morphing Time: Unleashing the Potential of
Multiple LLMs via Multi-objective Optimization}

\author{%
Bingdong~Li$^1$,%
~Zixiang~Di$^1$,%
~Yanting~Yang$^1$,%
~Hong~Qian$^1$,%
~Peng~Yang$^2$,\\%
~Hao~Hao$^{3,}$\thanks{Hao~Hao is the corresponding author.}~,%
~Ke~Tang$^2$,%
~Aimin~Zhou$^1$\\
$^{1}$East China Normal University\\
$^{2}$Southern University of Science and Technology\\
$^{3}$Shanghai Jiao Tong University\\
\texttt{bdli@cs.ecnu.edu.cn, \{51265901113, 51255901098\}@stu.ecnu.edu.cn,}\\
\texttt{hqian@cs.ecnu.edu.cn, yangp@sustech.edu.cn,}\\
\texttt{haohao@sjtu.edu.cn, tangk3@sustech.edu.cn, amzhou@cs.ecnu.edu.cn}
}

\maketitle

\begin{abstract}
In this paper, we introduce a novel approach for addressing the multi-objective optimization problem in large language model merging via black-box multi-objective optimization algorithms. The goal of model merging is to combine multiple models, each excelling in different tasks, into a single model that outperforms any of the individual source models. However, model merging faces two significant challenges: First, existing methods rely heavily on human knowledge or intuition. Second, it's difficult to obtain the great model merging configuration in limited evaluations. To address these challenges, we formalize model merging as a multi-objective optimization problem and propose an automated optimization approach named MM-MO. This method leverages multi-objective optimization algorithms to autonomously search for optimal merging configurations across various tasks, alleviating the need for human intervention. In MM-MO, a weak-to-strong method is employed to enhance the acquisition function, allowing previously evaluated superior configurations to guide the search for new ones. Meanwhile, Fisher information is applied to screen these configurations, increasing the possibility of identifying high-quality merging configuration. Additionally, we designed a sparsity metric as an additional optimization objective to enhance the model's generalization performance across different tasks. We conducted comprehensive experiments with other mainstream model merging methods, demonstrating that the proposed MM-MO algorithm is competitive and effective in achieving high-quality model merging.

\end{abstract}


\section{Introduction}
\IEEEPARstart{L}{arge} language models (LLMs)   have shown great
performance on tasks in various domains such as natural language processing
\cite{achiam2023gpt}, 
computer vision \cite{wang2024visionllm} etc.
With the great effort of world-wide contributors
in community, a large number of
general-purpose pre-trained and 
task-specific fine-tuned
language models have been proposed and made publicly available \cite{qwen, dou2024sailor, touvron2023llama, llama2, ivison2023camels, xu2023paradigm}.
However, LLM pre-training or fine-tuning is non-trivial and  requires a lot of effort and financial budget \cite{wang2023self}. 
Recently,
model merging (MM)
has
attracted many researchers' attention.
By combining multiple LLMs into a single model with
better performance and
adaptability on more tasks,
MM offers 
 a novel cost-efficient way of 
obtaining new powerful language models 
without requiring the collection of raw training data or expensive computation
\cite{yang2024model,akiba2024evolutionary,daheim2023model},
just like  the Power Rangers merge their Zords together to form a mighty Megazord 
\cite{Ries2024power}.
Ideally, MM is supposed to 
inherit and amplify the strengths from its source models
while ignoring their weaknesses.
Therefore, the obtained model
will be able to tackle the union set
of all the tasks where the source models
are pre-trained/fine-tuned
with better performance.
Yet this is achieved without training,
which saves a large amount of calculation/financial budget.
With the help of open-source toolkits such as mergekit
\cite{Goddard2024mergekit,Labonne2024MMmergekit},
 MM has become popular 
 for LLM developing and 
 shown great potential
on the Open LLM Leaderboard \cite{HuggingFace2024leader}.
 
However, model merging requires the model maker
to have profound knowledge or intuition \cite{akiba2024evolutionary}.
Alternatively, they may invest substantial time in conducting an extensive grid search to determine the configuration for model merging.
For example, DARE \cite{yu2024language} search the drop rate $p$ of each model in [0.1, 0.2, · · · , 0.9].
Since model merging involves not only determining the retention ratio of parameters but also setting the weight proportions of each model, the search space for a grid search expands substantially as the number of models increases. As a result, this approach becomes increasingly inefficient for identifying an optimal model merging configuration.
Automatically and efficiently discovering 
more capable MM recipes
is still in its infancy.
To the best of our knowledge, 
The most related work is  \cite{akiba2024evolutionary},
where
Akiba et al.
proposed to use 
evolutionary algorithms (EAs)
to generate powerful model merging recipes
which operates in both parameter space and data flow space.
However, the approach of using diverse task scores and optimizing the parameter space to explore the comprehensive potential of the final model has been largely overlooked.
 
MM is similar to ensemble learning \cite{yang2023survey} in that both methods aim to leverage the strengths of multiple source models to improve overall performance across diverse tasks.
In ensemble learning,
It is generally believed that 
diversity in an ensemble could be beneficial 
for  improving its performance  
\cite{brown2004diversity,tang2006analysis,yao2008evolving}.
One could infer that 
maintaining diversity  during MM
may also result in  powerful LLMs.
Having this in mind, we design a model merging approach via
multi-objective optimization
which
take into consideration multiple task-specific performance metrics simultaneously.

\begin{figure}[ht]
\begin{center}
\centerline{\includegraphics[width=0.35\textwidth]{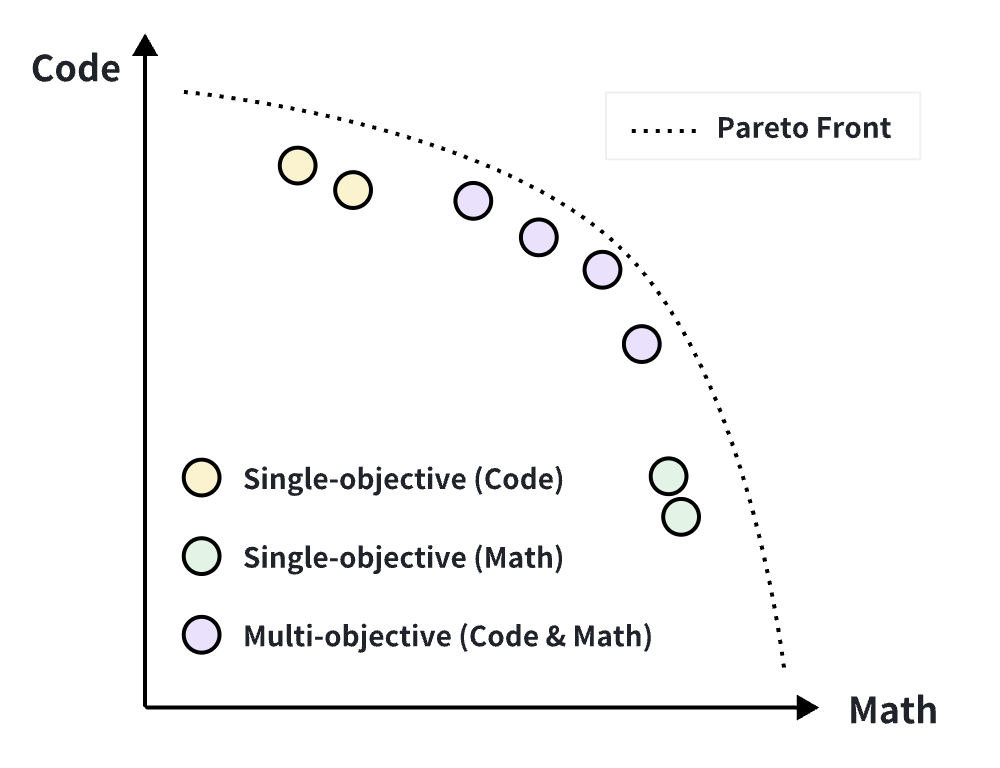}}
\vspace{-0.2cm}
\caption{Illustration of merged model performance differences when formalizing the model merging configuration search as a single-objective vs. multi-objective optimization problem. Multi-objective optimization achieves a more balanced performance across code and math capabilities.}
\label{fig:sop_mop}
\end{center}
\vspace{-0.4cm}
\end{figure}

In Fig 1, we illustrate the performance of merged models when the search for optimal model merging configurations is framed as either a single-objective or multi-objective optimization problem.  When focusing solely on the model’s code or math capabilities, single-objective optimized configurations tend to concentrate at the extremes, typically enhancing one capability at the expense of the other. In contrast, multi-objective optimized configurations achieve a more balanced performance, resulting in a merged model that attains satisfactory levels in both Code and Math abilities. Given that the goal of model merging is to combine multiple models, each excelling in different tasks, into a single model that performs well across various tasks, approaching the configuration search as a multi-objective optimization problem is a more effective and appropriate solution.

In this work, we proposed to leverage 
 {multi-objective optimization} for
automatic model merging. Specifically, we formulate
the problem of merging multiple models as a multi-objective optimization problem.
This automated model merging approach
is able to obtain a performance powerhouse that leverages the best qualities of each model, inheriting their respective strengths of all source models.
Furthermore, we validate the effectiveness of our MM-MO method through extensive evaluations across a diverse range of LLM evaluation tasks.
The contributions of this work can be summarized as follows: 
\\
(1) \textbf{Formalizing Model Merging as a Multi-Objective Optimization Problem:} We introduce a novel perspective by formulating model merging as a multi-objective optimization problem. This approach allows for the simultaneous consideration of multiple task-specific performance metrics, aiming to creating a comprehensive model that integrates the strengths of multiple source models across diverse tasks.
\\
(2) \textbf{Automated Model Merging via multi-objective optimization:} We introduce a novel approach to model merging that automates the search for better merging configurations using multi-objective black-box optimization algorithms, alleviating the reliance on human intuition and customized strategies.
\\
(3) \textbf{Enhanced Acquisition Strategy:} To achieve high-quality model merging configurations with limited evaluation iterations,
we introduced a weak-to-strong method to enhance acquisition function that allows previously evaluated superior configurations to guide new configurations' searching. Besides, we used Fisher information to screen these configurations to increase the chance of discovering superior merging configurations.
\\
(4) \textbf{Additional Optimization Objective:} In addition to task metrics, we have introduced a sparsity metric as another optimization objective to enhance the model's generalization performance across different tasks.

The remainder of this paper is organized as follows. Section \ref{sec:background} surveys the existing methods in model merging and multi-objective optimization. In Section \ref{sec:problem_formulation}, we formalize model merging as a multi-objective optimization problem, establishing the framework that underpins our approach. Section \ref{sec:method} introduces a novel multi-objective Bayesian optimization algorithm, MM-MO, tailored for model merging. In this section, we also present our improvements to the process of searching for new solutions and the design of new optimization objectives specifically for model merging scenarios. Section \ref{sec:experimental_setup} and  \ref{sec:experimental_results} presents the performance evaluation of the proposed method against related methods. Finally, we draw conclusions in Section \ref{sec:conclusion}.

\section{Background}
\label{sec:background}
  \subsection{Model Merging}

Generally speaking, there are four kinds of model merging (MM) methods.

One common approach is simply averaging the weights of different models, which has shown promising performance. Model soups \cite{wortsman2022model} achieved notable performance improvements on computer vision tasks. Task arithmetic \cite{ilharco2022editing} merges LLMs by performing arithmetic operations on task vectors (also known as delta parameters), which represent the differences between the weights of fine-tuned LLMs and the original pre-trained model. Fisher-Weighted Averaging \cite{matena2022merging} is based on the Laplace approximation, where each model's posterior is approximated as a Gaussian distribution, with its precision matrix corresponding to its Fisher information.

Spherical linear interpolation (SLERP) \cite{white2016sampling} can achieve smooth interpolation between two vectors and maintain directional consistency during model merging by calculating the angle between two vectors and determining the interpolation vector based on the interpolation factor and angle.

RegMean \cite{jin2022dataless} merges different models by minimizing the prediction differences between the merged model and the source models.

To address the interference between parameters of different models caused by redundant parameter values and disagreements on the sign of parameters across models, TIES-Merging \cite{yadav2024ties} introduces three novel steps when merging models: (1) resetting parameters with small delta parameters, (2) resolving sign conflicts, and (3) merging only the consistent parameters.

DARE \cite{yu2024language} performs model merging by sparsifying delta parameters of source supervised fine-tuned (SFT) models with Drop and Re-Scale operations before actually merging them.

Recently, Daheim et al. \cite{daheim2023model} connected the inaccuracy of weighted-averaging to mismatches in the gradients and proposed a new uncertainty-based scheme to improve performance by reducing this mismatch.

Our approach builds upon the combination of TIES-Merging and DARE, leveraging their strengths to create a robust model merging strategy. By integrating the conflict resolution techniques of TIES-Merging with the parameter sparsification methods of DARE, we aim to automate the search for better model merging configurations, ensuring that the final merged model maintains high performance without the degradation typically associated with parameter conflicts.

 \subsection{Multi-objective Optimization}
 
Without loss of generality,
a multi-objective optimization problem 
can be  stated as follows:
    \begin{equation}
            \begin{aligned} \label{mopdef}
    &\min \textbf{ } \boldsymbol{ f}(\boldsymbol{x})=(f_{1}(\boldsymbol{x}), f_{2}(\boldsymbol{x}),\ldots, f_M(\boldsymbol{x}))^{T}\\
    &\textrm{ }s.t.\textrm{ }  \boldsymbol{x} \in \mathcal{X}   
    \end{aligned},
  \end{equation} 
where 
$\boldsymbol{x}$ $=$ $(x_1,x_2,\ldots,x_d)$ is the decision vector, 
$\boldsymbol{f} (\cdot) $: $ \mathcal{X} \rightarrow \mathcal{T}$ is 
$m$
objective functions,
$ \mathcal{X} $ and $\mathcal{T}$ denote the (nonempty)  \textit{decision space} 
and  the \textit{objective space}, respectively.

In order to compare the quality of solutions of an MOP,
the concept of \textit{Pareto dominance} is introduced:
\begin{definition}[Pareto dominance~\cite{yu1974cone}]
	Given two solutions $\boldsymbol{a}$, $\boldsymbol{b} \in$ 
 $ \mathcal{X}$,
$\boldsymbol{a}$ is said to \textit{dominate} $\boldsymbol{b}$ (denoted as $\boldsymbol{a} \prec  \boldsymbol{b}$)
if and only if 
 $ \forall i \in \{1,2,...,m\}, f_{i}(\boldsymbol{a}) \leq f_{i}(\boldsymbol{b})$
and $\exists j \in \{1,2,...,m\}, f_{j}(\boldsymbol{a}) < f_{j}(\boldsymbol{b})$.
    A solution $\boldsymbol{a}^{\ast}  \in \mathcal{X}$
is  Pareto optimal
if no other solution $\boldsymbol{a}  \in \mathcal{X}$
  can dominate it.
    The  solution set consisting of all the Pareto optimal solutions is called the \textit{Pareto set} (PS): 
$PS$$=$$\{\boldsymbol{a}  \in \mathcal{X}| \forall \boldsymbol{b}  \in \mathcal{X},\boldsymbol{b} \not\prec  \boldsymbol{a}   \}$
and 
  the corresponding objective vector set of the PS is the \textit{Pareto front} (PF). 
\end{definition}

Multi-objective Optimization (MO) \cite{konak2006multi}
focuses on  approximating the PS, targeting at a solution set with good convergence and diversity in the objective space.
In this work, we utilize multi-objective optimization to address the complex challenge of model merging. By considering multiple objectives simultaneously, MO enables us to balance various performance metrics, such as different task-specific accuracies. This approach allows us to optimize the trade-offs between different performance criteria, leading to a more robust and effective merged model. Through this methodology, we aim to achieve a unified model that not only excels in specific tasks but also retains a broad and diverse set of capabilities from the individual models. This is particularly important in the context of LLMs, where different tasks may require different strengths and features from the base models. By leveraging MO, our method systematically explores the parameter space to identify better configurations, ensuring that the final model exhibits both high performance and comprehensive potential across various tasks.

\section{Problem Formulation}
\label{sec:problem_formulation}

\begin{figure}[ht]
\begin{center}
\centerline{\includegraphics[width=0.4\textwidth]{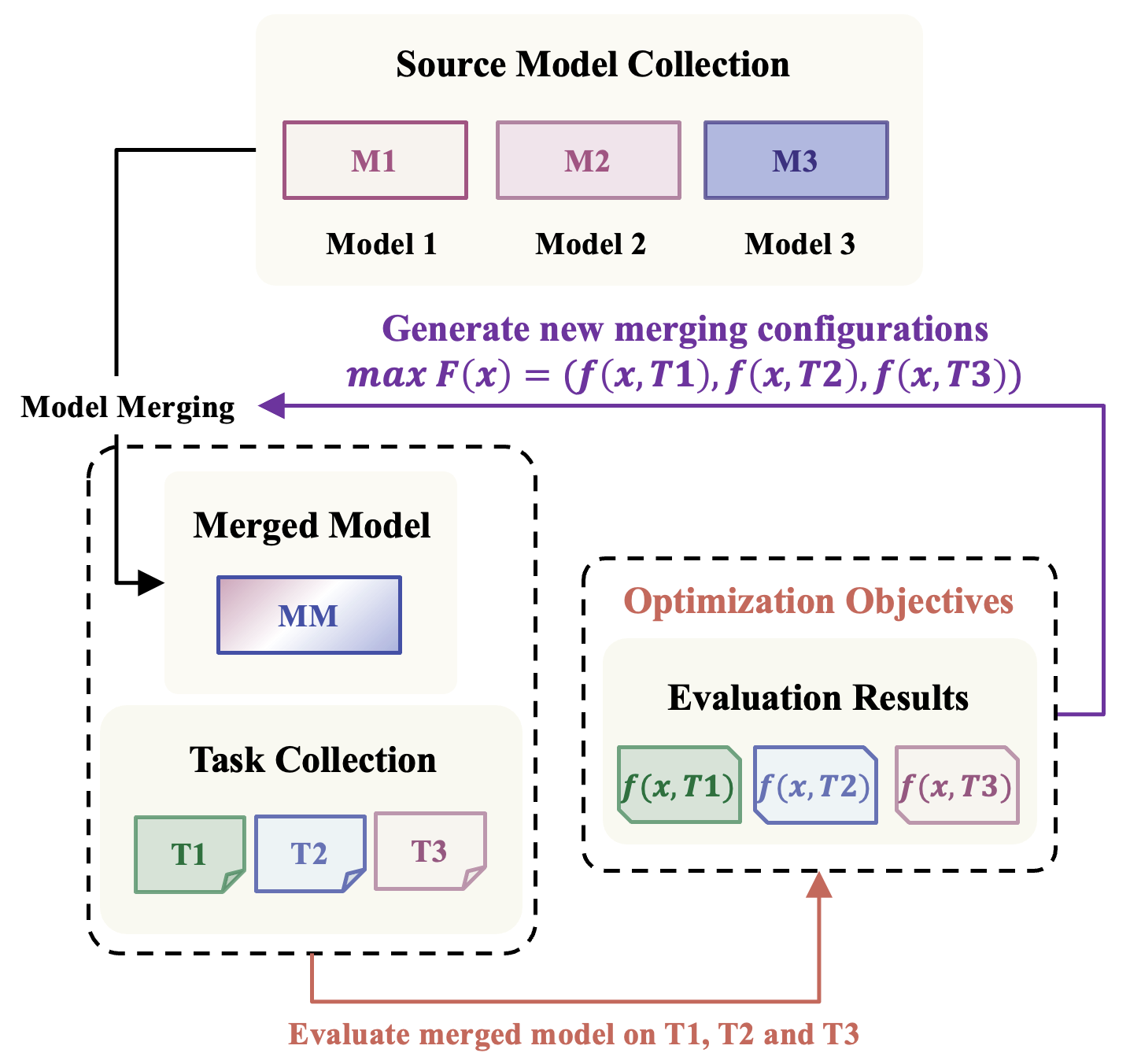}}
\vspace{-0.2cm}
\caption{An illustration of formalizing model merging as a multi-objective optimization problem, with three source models and three evaluation tasks.}
\label{fig:problem_formulation}
\end{center}
\vspace{-0.4cm}
\end{figure}

Here, we formalize model merging as a multi-objective optimization problem. The goal is to combine multiple homologous source models into a single, unified model that retains the strengths of each individual model and achieves superior performance across diverse tasks compared to any individual source model. This formalization allows us to automate the search for an optimal merging configuration, eliminating the reliance on human intuition and custom strategies.

The problem is framed as follows: Given a set of source models ${M_1, M_2, …, M_N}$ (In Fig. \ref{fig:problem_formulation}, $N = 3$), each of which is fine-tuned on distinct datasets but based on a common pre-trained backbone, our goal is to find a merging configuration that effectively balances task performance across multiple objectives. Each source model demonstrates strengths in different types of tasks, and the challenge is to derive a merged model that synthesizes these strengths for enhanced generalizability.

As illustrated in Fig. \ref{fig:problem_formulation}, the model merging process based on multi-objective optimization is divided into three main components:

1. \textbf{Configuration Search for Model Merging:} The process begins with the initialization of merging configurations $\boldsymbol{X} = {\boldsymbol{x}_1, \boldsymbol{x}_2, …}$, where each configuration $\boldsymbol{x}_i$ represents a specific configuration for merging the source models. For each configuration, a merged model is generated and evaluated based on its performance on various tasks, represented as optimization objectives $\boldsymbol{f}(\boldsymbol{x, T1}), \boldsymbol{f}(\boldsymbol{x, T2}), \dots, \boldsymbol{f}(\boldsymbol{x, TM})$, where each $\boldsymbol{f}(\boldsymbol{x, Tj})$ denotes the performance metric on task $\boldsymbol{Tj}$ for configuration $\boldsymbol{x}$.

2. \textbf{Objective Evaluation and Feedback:} For each merged model produced by a given configuration, task-specific performance metrics are calculated to assess its effectiveness across tasks. These metrics serve as feedback for refining configurations. This feedback loop enables the optimization algorithm to guide the search towards configurations that better capture the strengths of individual models, enhancing overall performance.

3. \textbf{Iterative Optimization Process:} The multi-objective optimization approach iteratively generates new merging configurations based on feedback from previous evaluations. This process repeats until a preset iteration limit is reached, ensuring that the final configuration achieves a balanced performance across all objectives. The outcome is a merging configuration that maximizes task performance while maintaining desirable generalization characteristics.

By formalizing model merging as a multi-objective optimization problem, we establish a structured foundation for an automated search for effective configurations, ultimately producing a robust, high-performing merged model.

In addition, formalizing the configuration search for model merging as a multi-objective optimization problem presents the following three challenges:

1. How to generate high-quality configurations? To address this challenge, we developed a weak-to-strong method to refine newly generated configurations (Subsection \ref{sec:weak-to-strong_method}), thereby increasing the probability of generating high-quality model merging configurations. 

2. How to identify superior configurations with fewer evaluation iterations? To address this challenge, we proposed a filtering approach based on Fisher information to select solutions for evaluation from each generation of candidates
(Subsection \ref{sec:fisher_information}). This method increases the possibility of discovering superior model merging configurations within the constraints of limited evaluation iterations.  

3. How to prevent the final merged model from overfitting to specific types of validation tasks? To address this challenge, we introduced an additional sparsity optimization objective, which effectively helps to prevent overfitting on particular evaluation tasks
(Subsection \ref{sec:sparity_optimization_objective}). This objective ensures that the merged model maintains robust overall performance and improved generalization capabilities. 

\section{Methodology}
\label{sec:method}
\subsection{Overview}
   
\begin{figure*}[ht]
\begin{center}
\centerline{\includegraphics[width=0.9\textwidth]{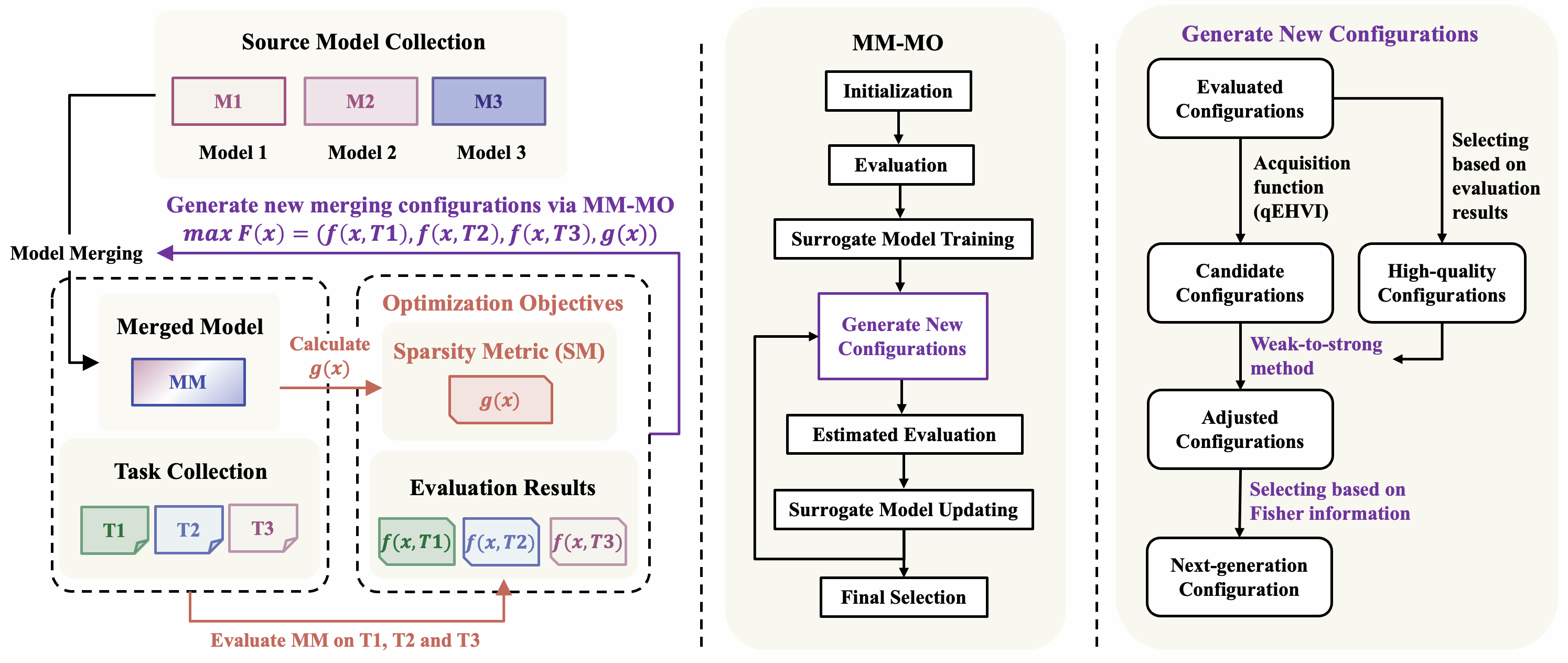}}
\vspace{-0.2cm}
\caption{An illustration of  automated model merging 
with multi-objective optimization (MM-MO).}
\label{fig:method_overview}
\end{center}
\vspace{-0.6cm}
\end{figure*}
 
Here, we provide an overview of our proposed automated model merging approach, MM-MO. Our approach is designed to merge multiple pre-trained models into a single, unified model that retains the strengths of each individual model and surpasses the performance of any single model. 
\textcolor{black}{
In Fig. \ref{fig:method_overview}, we show the overall process of model merging based on multi-objective optimization, which is divided into three parts:}

\textcolor{black}{
On the left is a schematic diagram of the automatic search process for model merging configurations. Our method starts with a set of source models, each obtained by performing SFT on different datasets using the same pre-trained backbone. These models have their own advantages for different task types. We merge these models to obtain a merged model, evaluate the performance of the merged model on different tasks as optimization objectives , and calculate the sparsity metric of the merged model as an additional optimization objective. For simplicity, we will denote $\boldsymbol{f}(\boldsymbol{x, Tj})$ later as $\boldsymbol{f_j}(\boldsymbol{x})$, $j\in\{1,...,M-1\}$. MM-MO generates new model merging configurations $\boldsymbol{X} = \{\boldsymbol{x}_1, \boldsymbol{x}_2, ...\}$ based on these objective values, allowing the creation of new merged models. This process is repeated until a preset evaluation iteration limit is reached, ultimately resulting in an optimal model merging configuration.}

\textcolor{black}{
In the middle is the complete process of using multi-objective Bayesian optimization to search for model merging configurations, detailed in Subsection \ref{sec:Model Merging via Multi-objective Optimization} and \ref{sec:pseudocode}.}

\textcolor{black}{
On the right is the step of generating new configurations. We enhanced the acquisition strategy of multi-objective Bayesian optimization by designing a weak-to-strong method to adjust candidate configurations, and we used Fisher information to select the next generation of model merging configurations, detailed in Subsection \ref{sec:Enhanced Acquisition Strategy}.}

\subsection{Model Merging via Multi-objective Optimization}
\label{sec:Model Merging via Multi-objective Optimization}
In this subsection, we detail the multi-objective optimization process employed for model merging. Our method leverages parallel multi-objective bayesian optimization (qEHVI) \cite{daulton2020differentiable}. This technique is an effective black-box optimization method designed to handle multiple objectives simultaneously, making it ideal for our model merging needs.

Our approach builds upon TIES-Merging with DARE \cite{yadav2024ties, yu2024language}, utilizing qEHVI to search for the optimal model merging configuration. Since TIES-Merging with DARE first applies sparsification to each source model before merging, our model merging configuration specifically includes the sparsity level of each source model (\texttt{Density}) and the weight proportion of each source model during merging (\texttt{Weight}). The integration of qEHVI allows for an automated and systematic exploration of the parameter space, identifying the best configurations that balance various performance metrics. Here are the detailed steps of our methodology.

1. \textbf{Initialization:} We start with a collection of pre-trained models, each fine-tuned on different tasks. These models and their configurations serve as the initial candidates for merging.

2. \textbf{Objective Definition:} We define multiple objectives that reflect the performance metrics we aim to optimize. 
\textcolor{black}{
Specifically, these objectives include task-specific accuracies, such as those from C-EVAL \cite{huang2023ceval} and GSM8K \cite{cobbe2021gsm8k}, as well as a sparsity metric $\boldsymbol{g}(\boldsymbol{x})$ used in our experiments.}
These metrics are crucial as they ensure the merged model performs well across different types of tasks.

3. \textbf{Surrogate Model Training:} qEHVI leverages surrogate models to approximate the objective functions. We train these surrogate models on the initial set of model configurations and their corresponding performance metrics. This step involves creating a probabilistic model that predicts the performance of unseen configurations.

4. \textbf{Acquisition Function Optimization:} The key component of qEHVI is the Expected Hypervolume Improvement (EHVI) acquisition function. This function guides the search by quantifying the expected improvement in the objective space. We optimize this acquisition function to select the most promising model configurations to evaluate next. Specifically, the hypervolume (HV) \cite{shang2020new} is a measure of the space dominated by a set of solutions in the objective space, bounded from below by a reference point. The HV indicator is defined as follows:
        \begin{equation}
            \textbf{HV}(S) = \Lambda(\{q \in \mathbb{R}^d| \exists p \in S:p \leq q\; \textbf{and}\; q \leq r \}),
            \label{hv}
        \end{equation}
where $S$ signifies a solution set, $r$ is identified as a reference vector set, and $\Lambda(\cdot)$ denotes the Lebesgue measure. Hypervolume improvement (HVI) measures the additional hypervolume that a new point contributes beyond the existing Pareto frontier.

In black-box optimization, exact function values at unobserved points are unknown, making direct HVI computation infeasible. Instead, the EHVI acquisition function uses the surrogate model's posterior distribution over the function values to compute the expected improvement. This is typically done using Monte Carlo (MC) integration to estimate EHVI \cite{emmerich2006single}, particularly when dealing with multiple candidates in parallel. The EHVI guides the search by identifying new points that are expected to provide the most significant improvement in the objective space. The more general parallel variant utilizing MC integration is given as follows:
\begin{equation}
    \begin{split}
        \alpha_{q\text{EHVI}}(\mathcal{X}_{\text{cand}}|\mathcal{P}) &\approx \hat{\alpha}_{q\text{EHVI}}(\mathcal{X}_{\text{cand}}|\mathcal{P}) \\
        &= \frac{1}{N}\sum_{t=1}^{N}\text{HVI}(\tilde{\boldsymbol{f}}_{t}(\mathcal{X}_{\text{cand}})|\mathcal{P}),
    \end{split}
    \label{qehvi}
\end{equation}
where $\tilde{f}_{t}\sim p(\boldsymbol{f}|\mathcal{D})$ for $t=1,...,N$ and $\mathcal{X}_{\mathrm{cand}}=\{x_{i}\}_{i=1}^{q}$ \cite{daulton2020differentiable}.
\textcolor{black}{
After obtaining candidate configurations using the acquisition function, we adjusted the candidate configurations using the weak-to-strong method. We then selected the configurations based on their Fisher information to ultimately obtain a new generation of model merging configurations. For details, please refer to Subsection \ref{sec:Enhanced Acquisition Strategy}.}

5. \textbf{Parallel Evaluation:} Unlike traditional methods, qEHVI supports parallel evaluations, which significantly speeds up the optimization process. In each iteration, we select a batch of model configurations and evaluate their performance on the defined objectives (e.g., C-EVAL and GSM8K accuracy).

6. \textbf{Updating the Surrogate Model:} After evaluating the selected configurations, we update the surrogate models with the new data. This iterative process continues, refining the surrogate models and improving the acquisition function's accuracy in identifying high-potential configurations.

7. \textbf{Convergence and Final Selection:} The optimization process continues until convergence criteria are met, such as a maximum number of iterations or negligible improvement in the objective space (in subsequent experiments, unless otherwise specified, the termination criterion is set as 30 real evaluations, i.e., FE=30). The final model merging configuration is selected based on the best trade-offs between the multiple performance metrics.

By using multi-objective Bayesian optimization method, our method systematically navigates the trade-offs between different task-specific accuracy, ensuring that the final merged model maintains high performance across various tasks. The iterative nature of this process allows for continuous improvement and fine-tuning, leading to a robust and high-performing unified model. This integration of qEHVI with TIES-Merging and DARE provides a powerful and automated solution for model merging, optimizing the balance between multiple performance criteria without relying on human intuition. This automated search for the optimal merging configuration ensures that the final model configuration surpasses the performance of any individual source model.

\subsection{Pseudocode of MM-MO}
\label{sec:pseudocode}
\begin{algorithm}[!htbp]
    \LinesNumbered
    \caption{The framework of MM-MO}
    \label{alg:mm-mo}
    \SetKwInOut{Input}{Input}\SetKwInOut{Output}{Output}
    \Input{
    number of iterations $K$, batch size $B$, number of initial configurations $N$}
    \Output{final configuration $\boldsymbol{X}_{Final}$}
    Initialize $N$ configurations $\boldsymbol{X}_0$ randomly\\
    Evaluate $\boldsymbol{Y}_0$ using objectives $\boldsymbol{f_1}(\boldsymbol{X}_0), \boldsymbol{f_2}(\boldsymbol{X}_0), ...,\boldsymbol{g}(\boldsymbol{X}_0)$\\
    \For{$k=0$ {\bfseries to} $K-1$}{
        Train surrogate model $GP_i^k$ based on $\{\boldsymbol{X}_k,\boldsymbol{Y}_k\}$ for each objectives.\\
        ${\boldsymbol{X}}_k^B \leftarrow$ Generate new configurations based on $GP^k$ and $\{\boldsymbol{X}_k, \boldsymbol{Y}_k\}$ \textbf{(Algorithm \ref{alg:gen_new_config})}\\
        Update $\boldsymbol{X}_{k+1} \leftarrow \boldsymbol{X}_k \cup {\boldsymbol{X}_k^B}$\\
        Evaluate $\boldsymbol{Y}_{k+1}$ using  $\boldsymbol{f_1}(\boldsymbol{X}_k^B), \boldsymbol{f_2}(\boldsymbol{X}_k^B), ...,\boldsymbol{g}(\boldsymbol{X}_k^B)$
    }
    $\boldsymbol{X}_{Final} \leftarrow$ Final selection from $\boldsymbol{X}_K$   
\end{algorithm}

\begin{algorithm}[!htbp]
    \LinesNumbered
    \caption{Generate new configuration}
    \label{alg:gen_new_config}
    \SetKwInOut{Input}{Input}\SetKwInOut{Output}{Output}
    \Input{evaluated configurations $\{\boldsymbol{X}, \boldsymbol{Y}\}$, acquisition function $\boldsymbol{qEHVI}$, surrogate model $GP$}
    \Output{next-generation configuration $\boldsymbol{X}_{New}$}
    $\boldsymbol{X}_{C} \leftarrow$ Generate candidate configurations based on $\boldsymbol{qEHVI}$ and $\{\boldsymbol{X}, \boldsymbol{Y}\} $\\
    $\boldsymbol{X}_{H} \leftarrow$ Select high-quality configurations based on $\{\boldsymbol{X}, \boldsymbol{Y}\}$\\
    $\boldsymbol{X}_{A} \leftarrow$ Adjust configurations by weak-to-strong method based on $\boldsymbol{X}_{C}$ and $\boldsymbol{X}_{H}$\\
    $\boldsymbol{X}_{New} \leftarrow$ Select the next-generation configuration by Fisher information based on $\boldsymbol{X}_{A}$
\end{algorithm}

In this section, we present the pseudocode for the MM-MO framework (Algorithm \ref{alg:mm-mo}) and for generating new configurations (Algorithm \ref{alg:gen_new_config}). The basic steps of the overall MM-MO process are as follows:

\begin{enumerate}
    \item 
 \textbf{Initialization Phase}:
The process begins by randomly generating an initial set of $N$ configurations, denoted as $\boldsymbol{X}_0$, where $N$ defaults to 10.
These configurations are evaluated using the evaluation function $\boldsymbol{f}(\boldsymbol{x})$ and the sparsity metric calculating function $\boldsymbol{g}(\boldsymbol{x})$. The results are combined into an initial set of objective values, $\boldsymbol{Y}_0$, which includes both evaluation results and sparsity metrics.
 \item 
\textbf{Iterative Optimization Phase}:
The algorithm iterates $K$ times, where $K$ defaults to 4. In each iteration, the current set of configurations $\boldsymbol{X}_k$ and their associated objective values $\boldsymbol{Y}_k$ are used to train surrogate models (Gaussian Processes, GP) for each objective, which is a commonly used surrogate model in Bayesian optimization.
Based on these surrogate models and the current evaluated configurations, a batch of new candidate configurations ${\boldsymbol{X}}_k^B$ is generated. We set the batch size to 5 in our experiment.
The newly generated configurations are added to the existing set, and their objective and sparsity metrics are evaluated. This updates both the configuration set and their objective values for the next iteration.
 \item 
\textbf{ Final Selection Phase}:
After completing all iterations, the configuration with the highest objective values is selected as the final configuration $\boldsymbol{X}_{Final}$, which serves as the output of the algorithm.
\end{enumerate} 

Additionally, when   generating  new configurations, the algorithm first generates candidate configurations based on the acquisition function (e.g., qEHVI \cite{daulton2020differentiable}) and the previously evaluated configurations. It then selects high-quality configurations based on objective values, applies a weak-to-strong adjustment method, and finally selects half of the configurations according to the Fisher information for the next generation. This approach can increase the chances of discovering excellent model merging configurations within limited number of evaluations.

\subsection{Enhanced Acquisition Strategy}
\label{sec:Enhanced Acquisition Strategy}
\textcolor{black}{
To achieve high-quality model merging configurations within limited number of evaluation iterations, we have further improved the acquisition function of multi-objective Bayesian optimization for the model merging scenario.}

\textcolor{black}{
Specifically, we designed a weak-to-strong method, which adjusts the candidate configurations obtained using the acquisition function based on already evaluated superior configurations. In addition, we employ Fisher information to perform environment selection on the adjusted solutions, thereby further enhancing the chance of discovering superior model merging configurations.}

\textcolor{black}{
The details of these two methods will be introduced in the following two sections.}

\subsubsection{\textbf{Weak-to-strong Method}}
\label{sec:weak-to-strong_method}
The Weak-to-Strong method has previously been utilized in aligning LLMs. EXPO \cite{zheng2024weak} allows for the direct acquisition of a better-aligned model by extrapolating from the weights of an aligned model and its initial SFT checkpoint, eliminating the need for additional training. Inspired by this methodology, we developed a weak-to-strong method \textcolor{black}{to improve the process of searching for new configurations} in our model merging scenario.

The weak-to-strong method operates as follows:

\begin{enumerate}
    \item Selection of Superior Configurations: From the already evaluated configurations, we select a small number (set to 5 in our experiments) of the top-performing configurations.
    \item Differential Evolution: We apply Differential Evolution (DE) \cite{das2010differential} to the parameters of these selected configurations. DE is a population-based optimization algorithm that creates new candidate solutions by combining the parameters of existing ones, ensuring diversity and exploration of the search space. Specifically, DE generates new candidate solutions using the following formula:
\begin{equation}
\mathbf{v}_i^{(k+1)}=\mathbf{x}_{r_1}^{(k)}+F\cdot(\mathbf{x}_{r_2}^{(k)}-\mathbf{x}_{r_3}^{(k)})
\label{de}
\end{equation}

where $\mathbf{v}_i^{(k+1)}$ is the new solution vector for the $i$ -th individual at generation $k+1$. $\mathbf{x}_{r_1}^{(k)}$, $\mathbf{x}_{r_2}^{(k)}$, and $\mathbf{x}_{r_3}^{(k)}$ are randomly selected distinct individuals from the current population, and $r_1\neq r_2\neq r_3\neq i$. The differential mutation factor $F$ is a scaling factor typically between 0 and 1.

\item Stochastic Perturbation: To introduce an element of randomness and avoid local optima, we apply a stochastic perturbation with a probability of 50\%. This perturbation replaces some of the parameters in the candidate configurations with those obtained from the DE process, encouraging exploration beyond the immediate neighborhood of the superior configurations.
\end{enumerate}

This method helps in refining the acquisition function's output by focusing the search around promising areas in the configuration space, thereby increasing the probability of finding high-quality model merging configurations.

\subsubsection{\textbf{Fisher Information}}
\label{sec:fisher_information}
In the context of our model merging scenario, the purpose is to identify high-quality configurations within limited number of evaluation iterations. To achieve this, we need to refine our surrogate model efficiently. One way to do this is by carefully selecting which configurations to evaluate based on their potential to improve the surrogate model's accuracy. Fisher information \cite{ly2017tutorial} serves as a pivotal tool in this selection process. By prioritizing configurations with lower Fisher information for evaluation, we can enhance the precision of our surrogate model more efficiently.

Fisher information is a measure of the amount of information that an observable variable provides about an unknown parameter, thereby quantifying the expected information gain from data regarding parameter estimation. Mathematically, for a parameter $\theta$ and a probability density function $p(x;\theta)$ , the Fisher information $I(\theta)$ is expressed as:
\begin{equation}
    I(\theta)=-\mathbb{E}\left[\frac{\partial^2\log p(x;\theta)}{\partial\theta^2}\right] = \mathbb{E} \left[ \left( \frac{\partial \log p(x; \theta)}{\partial \theta} \right)^2 \right].
\end{equation}
In our methodology, the Fisher information for each candidate configuration is calculated using the current surrogate model, which is modeled as a Gaussian process \cite{balandat2020botorch}, a commonly used surrogate model in Bayesian optimization. The detailed process is as follows:

\begin{enumerate}
\item Posterior Predictive Calculation: For each candidate configuration, the Gaussian process surrogate model is employed to compute the posterior predictive mean and variance.

\item Fisher Information Approximation: The variance of the posterior distribution is utilized to approximate the Fisher information for each configuration. Configurations with lower Fisher information indicate regions where the surrogate model's predictions are less certain.

\item Environment Selection: Configurations with lower Fisher information are prioritized for evaluation. This ensures that configurations contributing more significantly to the model's uncertainty are evaluated first, thereby enhancing the surrogate model's overall accuracy.
\end{enumerate}

By incorporating Fisher information into the environment selection process, we aim to improve the efficiency of the evaluation process, enhances the accuracy of the surrogate model more rapidly, and increases the possibility of discovering superior model merging configurations within the constraints of limited evaluation iterations.

\subsection{Sparsity Optimization Objective}
\label{sec:sparity_optimization_objective}
In addition to evaluating the performance of each merged model across various tasks, we introduce an additional optimization objective: a sparsity metric $\boldsymbol{g}(\boldsymbol{x})$ This metric is designed to quantify the sparsity of the delta parameters between the merged model and the base model. Specifically, $\boldsymbol{g}(\boldsymbol{x})$ is computed by taking the difference between the Merged Model (MM) and the Base Model to obtain the Delta Parameters, and then calculating the L1-Norm of all Delta Parameters. The sparsity metric $\boldsymbol{g}(\boldsymbol{x})$ can be expressed mathematically as:
\begin{equation}
f_M(\boldsymbol{x}) = \boldsymbol{g}(\boldsymbol{x})=\sum_{i=1}^n\|\Delta\theta_i\|_1,
\end{equation}
where $\Delta\theta_i$ represents the delta parameters.

The purpose of incorporating this additional optimization objective is to prevent the final merged model from overfitting to specific evaluation tasks, thereby ensuring the merged model exhibits better overall performance and generalization capabilities.

The rationale behind designing this sparsity metric is supported by the findings of Yu et al. in their research on Dare \cite{yu2024language}, which demonstrated that for LLMs, removing up to 90\% of the delta parameters does not significantly degrade performance. In some cases, the drop rate $p$ can even reach 99\%, indicating that the supervised fine-tuning introduces substantial redundancy in the model parameters. This suggests that only minor modifications to the base model parameters are needed to unlock the model's potential and achieve optimal results. 

Based on these insights, we calculate the $L^1\text{-norm}$ of all delta parameters and use it as an additional optimization objective. This sparsity objective acts as a regularization term during the model merging parameter configuration search, aiming to minimize the redundancy of delta parameters. By doing so, it helps to prevent the final merged model from overfitting to specific types of evaluation tasks and ensures that the merged model maintains robust overall performance and improved generalization capabilities.

\section{Experimental Setup}
\label{sec:experimental_setup}
\subsection{Source Models}
To develop a model with strong comprehensive abilities across various disciplines, we applied our proposed model merging method to a set of source models, including Qwen1.5-7B-Chat \cite{qwen}, Liberated-Qwen1.5-7B \cite{Liberated-Qwen1.5-7B}, and firefly-qwen1.5-en-7B \cite{firefly-qwen1.5-en-7b}. Besides these Qwen1.5-7B models, we also experimented with models of different sizes and pre-trained backbones (Qwen2-1.5B\cite{qwen2} and Llama-2-13b\cite{llama2}), including Qwen2-1.5B-Instruct \cite{qwen2}, SauerkrautLM-1.5b \cite{SauerkrautLM-1.5b}, Qwen2-1.5B-Ita \cite{Qwen2-1.5B-Ita}, WizardLM-13B \cite{wizardlm}, WizardMath-13B \cite{wizardmath} and llama-2-13b-code-alpaca \cite{codealpaca}. Please see Table \ref{tab:backbones} for their versions and correspondences with pre-trained backbones.

\begin{table*}[h!]
    \centering
    \caption{Versions and correspondences with pre-trained backbones of SFT LLMs.}
    \begin{tabular}{c c c c}
        \toprule
        \textbf{Parameter Scale} & \textbf{SFT LLMs} & \textbf{URL} & \textbf{Pre-Trained Backbones} \\
        \midrule
        \multirow{3}{*}{7B} & Qwen1.5-7B-Chat \cite{qwen} & \url{https://huggingface.co/Qwen/Qwen1.5-7B-Chat} & \multirow{3}{*}{Qwen1.5-7B \cite{qwen}} \\
        & Liberated-Qwen1.5-7B \cite{Liberated-Qwen1.5-7B} & \url{https://huggingface.co/abacusai/Liberated-Qwen1.5-7B} & \\
        & firefly-qwen1.5-en-7B \cite{firefly-qwen1.5-en-7b} & \url{https://huggingface.co/YeungNLP/firefly-qwen1.5-en-7b} & \\
        \hline
        \multirow{3}{*}{1.5B} & Qwen2-1.5B-Instruct \cite{qwen2} & \url{https://huggingface.co/Qwen/Qwen2-1.5B-Instruct} & \multirow{3}{*}{Qwen2-1.5B \cite{qwen2}} \\
        & SauerkrautLM-1.5b \cite{SauerkrautLM-1.5b} & \url{https://huggingface.co/VAGOsolutions/SauerkrautLM-1.5b} & \\
        & Qwen2-1.5B-Ita \cite{Qwen2-1.5B-Ita} & \url{https://huggingface.co/DeepMount00/Qwen2-1.5B-Ita} & \\
        \hline
        \multirow{3}{*}{13B} & WizardLM-13B \cite{wizardlm} & \url{https://huggingface.co/WizardLM/WizardLM-13B-V1.2} & \multirow{3}{*}{Llama-2-13b \cite{llama2}} \\
        & WizardMath-13B \cite{wizardmath} & \url{https://huggingface.co/WizardLM/WizardMath-13B-V1.0} & \\
        & llama-2-13b-code-alpaca \cite{codealpaca} & \url{https://huggingface.co/layoric/llama-2-13b-code-alpaca} & \\
        \bottomrule
    \end{tabular}
    \label{tab:backbones}
\end{table*}

\subsection{Datasets}

For our study, we divided the usage of datasets into two distinct phases: configuration search and final model evaluation.

\textcolor{black}{
In the \textbf{configuration search} phase, we aimed to optimize the configurations of MM without overfitting to the test datasets. For assessing the model’s general abilities, we used the C-EVAL \cite{huang2023ceval} Validation Set, a Multi-Level Multi-Discipline Chinese Evaluation Suite, to ensure the model's comprehensive abilities across different disciplines were appropriately calibrated. Notably, when merging Llama-series models, we replaced C-EVAL with MMLU \cite{hendrycks2020measuring}, considering the Llama models’ stronger proficiency in answering English-based questions. MMLU, which also evaluates general model capabilities, provided a more suitable benchmark for these models. We present the impact of using different language validation sets in Appendix B-E. Additionally, to evaluate and optimize the mathematical reasoning abilities of our model, we sampled 1\% of the GSM8K \cite{cobbe2021gsm8k} Training Set. Since GSM8K does not have a predefined test set, the validation set is typically used for testing. To avoid data leakage, we selected a small, randomly chosen subset from the training set.}

\textcolor{black}{
For the \textbf{final evaluation} of our models, we employed a range
of datasets tailored to assess various capabilities, including C-EVAL \cite{huang2023ceval}, MMLU \cite{hendrycks2020measuring}, GSM8K \cite{cobbe2021gsm8k}, HellaSwag \cite{zellers2019hellaswag}, HumanEval \cite{chen2021evaluating}, MBPP \cite{austin2021program}, WinoGrande \cite{sakaguchi2021winogrande} and BIG-Bench Hard (BBH) \cite{suzgun2022challenging}.}

Among these, C-EVAL and MMLU are multidisciplinary examinations that evaluate the model's comprehensive abilities; GSM8K assesses the model's mathematical reasoning abilities; HellaSwag evaluates the model's commonsense reasoning abilities; HumanEval and MBPP assess the model's code generation abilities; WinoGrande evaluates the model's language understanding abilities; and BBH assesses the model's comprehensive reasoning abilities. 
All detailed descriptions of the datasets are as follows:

\begin{itemize}
\item{C-EVAL \cite{huang2023ceval} and MMLU \cite{hendrycks2020measuring}.}
C-EVAL and MMLU are multidisciplinary examinations assessing a model's general knowledge across various subjects such as humanities, sciences, and social studies.

\item{GSM8K \cite{cobbe2021gsm8k}.}
The GSM8K dataset focuses on evaluating mathematical reasoning abilities through a collection of grade school math problems, testing the model's problem-solving skills in mathematics.

\item{HumanEval \cite{chen2021evaluating} and MBPP \cite{austin2021program}.}
HumanEval and MBPP are used to assess code generation abilities. HumanEval contains code snippets and problem descriptions, while MBPP includes programming problems to test code generation accuracy.

\item{HellaSwag \cite{zellers2019hellaswag}.}
HellaSwag assesses commonsense reasoning by presenting multiple-choice questions that require the model to complete scenarios based on everyday common sense.

\item{WinoGrande \cite{sakaguchi2021winogrande}.}
WinoGrande evaluates language understanding by presenting sentences with ambiguous pronouns, requiring the model to identify the correct antecedents.

\item{BIG-Bench Hard (BBH) \cite{suzgun2022challenging}.}
BBH assesses comprehensive reasoning abilities with challenging questions that test advanced reasoning and problem-solving skills.
\end{itemize}

By utilizing these datasets, we ensured a comprehensive and balanced evaluation of models across a wide range of abilities and tasks.

\subsection{Baselines}
We have compared MM-MO with 5 state-of-
the-art and classical merging methods, including Linear (Model Soups) \cite{wortsman2022model}, Task Arithmetic \cite{ilharco2022editing}, TIES \cite{yadav2024ties}, DARE \cite{yu2024language} and Model Breadcrumbs \cite{davari2023model}. 
All detailed descriptions of the merging algorithms used for comparison are as follows:

\begin{itemize}
\item{Linear (Model Soups) \cite{wortsman2022model}.}
Model Soups improves model accuracy by averaging the weights of multiple models fine-tuned with different hyperparameters. This method enhances robustness without additional inference or memory costs, making it effective for large pre-trained models in tasks like image classification and natural language processing.

\item{Task Arithmetic \cite{ilharco2022editing}.}
Task Arithmetic uses task vectors, which represent directions in the weight space of a pre-trained model, to modify and combine model behaviors. By performing arithmetic operations on these vectors, the model's performance can be steered across multiple tasks efficiently and effectively, without requiring additional training data for each task.

\item{TIES \cite{yadav2024ties}.}
TIES (TRIM, ELECT SIGN \& MERGE) addresses parameter interference in model merging by resetting minimally changed parameters, resolving sign conflicts, and merging only aligned parameters. This method outperforms existing techniques in various settings, maintaining valuable information and ensuring better performance in multi-task models.

\item{DARE \cite{yu2024language}.}
DARE (Drop And Rescale) sparsifies delta parameters by setting most to zero and rescaling the rest. This approach mitigates parameter interference when merging multiple fine-tuned models into one, enhancing capabilities without retraining. DARE proves particularly effective for large-scale language models, often surpassing the performance of individual source models.

\item{Model Breadcrumbs \cite{davari2023model}.}
Model Breadcrumbs enhance task performance by defining a sparse set of weights that create a trajectory within the weight space of a pre-trained model. This method improves efficiency and performance across multiple tasks without needing hyperparameter tuning.
\end{itemize}

Details of Parameter Settings for Baseline:
We use the grid search method to determine the model merging configurations for the baselines, which is a commonly adopted practice in current model merging studies \cite{yu2024language,  davari2023model}. Table \ref{tab:grid_search_ranges} presents the parameter search ranges for each model merging method. We search the parameters of these baseline methods and select the optimal setting with the best performance. This approach aligns with previous work on model merging, as demonstrated in the DARE \cite{yu2024language} paper. Specifically, \texttt{Weight} represents the weight proportion of each source model during merging, and \texttt{Density} denotes the sparsity level of each source model.

\begin{table}[h!]
    \centering
    \caption{Grid searched ranges of parameters of each baseline.}
    \vspace{-0.2cm}
    \resizebox{0.45\textwidth}{!}{
    \begin{tabular}{l l}
        \toprule
        \textbf{Model Merging Methods} & \textbf{Grid Search Ranges} \\
        \midrule
        Linear (Model Soups) & 
        \texttt{Weight}: [0.1, 0.3, 0.5, 0.7, 0.9, 1.0] \\
        \midrule
        Task Arithmetic & 
        \texttt{Weight}: [0.1, 0.3, 0.5, 0.7, 0.9, 1.0] \\
        \midrule
        TIES & 
        \texttt{Weight}: [0.1, 0.3, 0.5, 0.7, 0.9, 1.0] \\
        & \texttt{Density}: [0.1, 0.3, 0.5, 0.7, 0.9, 1.0] \\
        \midrule
        DARE & 
        \texttt{Weight}: [0.1, 0.3, 0.5, 0.7, 0.9, 1.0] \\
        & \texttt{Density}: [0.1, 0.3, 0.5, 0.7, 0.9, 1.0] \\
        \midrule
        Model Breadcrumbs & 
        \texttt{Weight}: [0.1, 0.3, 0.5, 0.7, 0.9, 1.0] \\
        & \texttt{Density}: [0.1, 0.3, 0.5, 0.7, 0.9, 1.0] \\
        \bottomrule
    \end{tabular}
    }
    \label{tab:grid_search_ranges}
\end{table}

\subsection{Evaluation Metrics}
We calculate 5-shot accuracy for C-EVAL, 5-shot weighted accuracy for MMLU, 0-shot and 5-shot accuracy for GSM8K (0-shot for 7B model and 5-shot for smaller 1.5B model), 0-shot accuracy for HellaSwag, pass@1 for HumanEval and MBPP, 0-shot accuracy for WinoGrande, and 3-shot accuracy for BBH.

\section{Experimental Results and Analysis}
\label{sec:experimental_results}

The experimental sections of this paper are structured as follows.

In the main paper, we evaluate the general performance of MM-MO and provide detailed results on the BBH dataset (Subsection \ref{sec:General_Performance} and \ref{sec:Comparison_on_BBH_Dataset}). Next, we compare MM-MO with the SOTA evolutionary algorithm-based model merging method (Subsection \ref{sec:MM-MO_vs_EMM}). We also investigate the effects of different acquisition functions and varying model sizes and backbones for source models on MM-MO (Subsection \ref{sec:MM-MO_acquistion_function} and \ref{sec:MM-MO_model_size}). Parameter sensitivity experiments are conducted to assess the impact of the number of evaluations (Subsection \ref{sec:parameter_sensitivity}), followed by an ablation study of all modules within MM-MO (Subsection \ref{sec:ablation_study}). 

In the Appendix, we provide further supplementary experiments, including detailed results on the C-EVAL dataset (Appendix B-A), case studies on Mathematical Reasoning and Coding (Appendix B-B and B-C), an evaluation of MM-MO with a larger number of source models (Appendix B-D), and a comparative analysis of the validation sets used (Appendix B-E), offering a comprehensive evaluation of MM-MO’s performance and flexibility.

\subsection{General Performance}
\label{sec:General_Performance}

\begin{table*}[h!]
    \centering
    \caption{Performance Comparison of different merging methods and single models. (Model size: 7B params)}
    \vspace{-0.2cm}
    \resizebox{\textwidth}{!}{
    \begin{tabular}{l l c c c c c c c c}
        \toprule
        \textbf{Merging Method} & \textbf{Models} & \textbf{Average Score} & \textbf{C-EVAL} & \textbf{GSM8K} & \textbf{HellaSwag} & \textbf{HumanEval} & \textbf{MBPP} & \textbf{MMLU} & \textbf{WinoGrande} \\
        \midrule
        Single Model 1 & Qwen1.5-7B-Chat & 56.46 & 68.7 & 54.59 & 68.13 & 46.95 & 34.20 & 60.06 & 62.59 \\
        Single Model 2 & Liberated-Qwen1.5-7B & 57.29 & 69.7 & 53.30 & 71.02 & 48.78 & 38.80 & 58.84 & 60.62 \\
        Single Model 3 & firefly-qwen1.5-en-7b & 51.32 & 70.0 & 49.81 & 65.69 & 33.54 & 28.40 & 51.66 & 60.14 \\
        \midrule
        Linear (Model Soup) & Single Model 1 + 2 + 3 & 58.88 & 71.1 & 54.89 & 72.72 & 50.61 & 39.80 & 60.80 & 62.27 \\
        Task Arithmetic & Single Model 1 + 2 + 3 & 56.67 & 70.1 & 55.50 & 69.54 & 46.34 & 37.80 & 55.04 & 62.35 \\
        Dare + Task Arithmetic & Single Model 1 + 2 + 3 & 58.38 & 69.8 & 55.27 & 69.97 & 51.22 & 39.40 & 60.23 & 62.75 \\
        TIES & Single Model 1 + 2 + 3 & 53.32 & 65.7 & 53.15 & 69.58 & 34.76 & 29.00 & 58.05 & 62.98 \\
        DARE + TIES & Single Model 1 + 2 + 3 & 57.78 & 69.5 & 55.72 & 69.23 & 49.39 & 37.80 & 60.06 & 62.75 \\
        Model Breadcrumbs & Single Model 1 + 2 + 3 & 58.56 & 70.3 & 56.10 & 70.91 & 49.39 & 40.60 & 60.47 & 62.12 \\
        Model Breadcrumbs + TIES & Single Model 1 + 2 + 3 & 58.41 & 70.2 & 55.72 & 70.59 & 50.00 & 39.80 & 60.35 & 62.19 \\
        \midrule
        DARE + TIES w/ MM-MO (Ours) & Single Model 1 + 2 + 3 & \textbf{60.97} & \textbf{71.9} & \textbf{57.77} & \textbf{74.44} & \textbf{55.49} & \textbf{42.20} & \textbf{60.81} & \textbf{64.17} \\
        \bottomrule
    \end{tabular}
    }
    \label{tab:general_performance}
\end{table*}

In the general experiment, we compared the performance of different model merging methods against individual models. The table presents the average scores and specific performances of each single model and their merged versions across various tasks.

\textbf{Performance Analysis of Single Models.}
These three SFT models exhibited varying performances across the tasks. Specifically:

Single Model 1 performed best in the C-EVAL task (68.7) but had relatively lower scores in HumanEval (46.95) and MBPP (34.20).

Single Model 2 showed strong performance in HellaSwag (71.02) and MBPP (38.80) but underperformed in GSM8K (53.30) compared to Single Model 1.

Single Model 3 excelled in C-EVAL (70.0) but performed poorly in HumanEval (33.54) and MBPP (28.40).

These differences indicate that each single model has its own strengths and weaknesses across different tasks.

\textbf{Comparison of Model Merging Strategies.}
Existing model merging methods surpassed individual models in certain tasks. For instance, the DARE + TIES method achieved higher scores in GSM8K (55.72) and HumanEval (49.39) than any single model. However, these methods do not guarantee significant improvements across all tasks. For example:

Linear merging performed well in C-EVAL (71.1) and HellaSwag (72.72), but its performance in WinoGrande (62.27) was lower than that of Single Model 1 and Single Model 3.

Task Arithmetic merging showed good results in GSM8K (55.50) but had an overall average score (56.67) lower than that of Single Model 2 (57.29).

DARE + Task Arithmetic improved scores in HumanEval (51.22) and MBPP (39.40) but did not surpass Single Models 1 and 2 in C-EVAL (69.8) and WinoGrande (62.75).

TIES method underperformed compared to individual models in some tasks, such as the average score (53.32) and HumanEval (34.76).

Breadcrumbs method generally performed well, with an average score of 58.56 and notable improvements in GSM8K (56.10) and MBPP (40.60), but did not perform as well as DARE + Task Arithmetic in HumanEval (49.39 vs. 51.22).

These results indicate that current methods do not ensure consistent improvements across all tasks. Existing model merging algorithms still have significant room for improvement in handling parameter conflicts and weight allocation during model merging. Therefore, we believe that further exploration of merging configurations can more effectively enhance the overall performance of merged models, achieving more comprehensive performance improvements.

\textbf{Performance Analysis of Our method (MM-MO).}
MM-MO demonstrated the best overall performance, achieving the highest average score of 60.97 and it excelled in all evaluated tasks. This method's superior performance across all tasks validates its effectiveness in optimizing model merging.

Furthermore, it is noteworthy that although we only used the accuracies of the C-EVAL and GSM8K datasets as optimization targets during the search configuration process, the final results show that the merged model obtained using MM-MO not only exhibits significant performance improvements on the C-EVAL and GSM8K test sets but also demonstrates noticeable enhancements in other types of tasks. This indicates that using MM-MO for model merging not only leverages the strengths of individual models in specific tasks but also enhances the overall potential of the model. It provides a reliable solution for integrating different single models into a high-performance and comprehensive model.

In summary, although the individual models show varying performance across different tasks, effective model merging strategies can substantially enhance overall performance. Our MM-MO method consistently achieved the best results in all evaluated tasks, validating the significant potential of merged models in tackling different tasks.

\subsection{Comparison on BBH Dataset}
\label{sec:Comparison_on_BBH_Dataset}

\begin{table}[h!]
    \centering
    \caption{Average accuracy over 27 BBH tasks.}
    \vspace{-0.2cm}
    \resizebox{0.45\textwidth}{!}{
    \begin{tabular}{l l c}
        \toprule
        \textbf{Merging Method} & \textbf{Models} & \textbf{BBH} \\
        \midrule
        Single Model 1 & Qwen1.5-7B-Chat & 38.02 \\
        Single Model 2 & Liberated-Qwen1.5-7B & 42.42 \\
        Single Model 3 & firefly-qwen1.5-en-7b & 28.41 \\
        \midrule
        DARE + TIES w/ MM-MO (Ours) & Single Model 1 + 2 + 3 & \textbf{47.72} \\
        \bottomrule
    \end{tabular}
    }
    \vspace{-0.2cm}
    \label{tab:bbh_performance}
\end{table}

\begin{figure}[ht]
\begin{center}
\centerline{\includegraphics[width=0.35\textwidth]{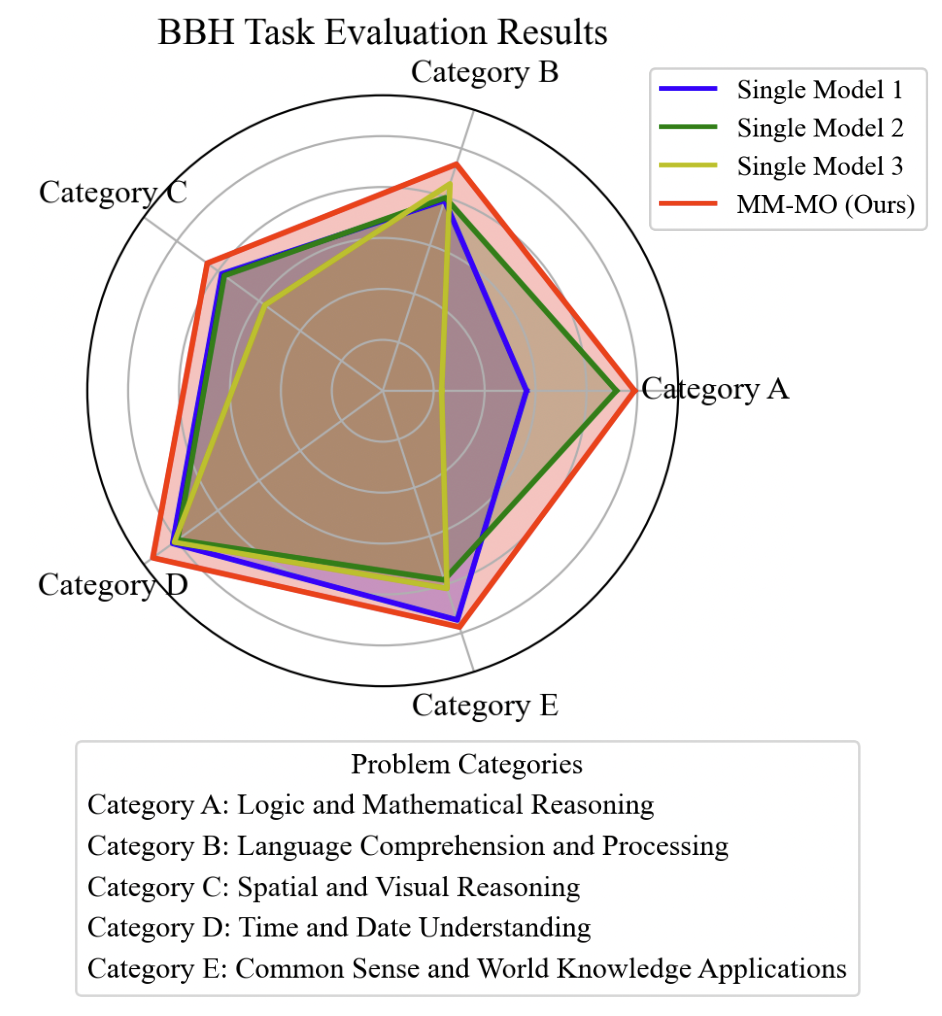}}
\vspace{-0.2cm}
\caption{Performance of single and merged LLMs on different categories of the BBH dataset.}
\label{fig:bbh_radar}
\end{center}
\vspace{-0.4cm}
\end{figure}

\begin{table}[h!]
    \centering
    \caption{Categorization of BBH tasks.}
    \vspace{-0.2cm}
    \resizebox{0.48\textwidth}{!}{
    \begin{tabular}{l l}
        \toprule
        \textbf{Category} & \textbf{Tasks} \\
        \midrule
        Logical and Mathematical Reasoning & 
        logical\_deduction\_three\_objects \\
        & logical\_deduction\_five\_objects \\
        & logical\_deduction\_seven\_objects \\
        & boolean\_expressions \\
        & formal\_fallacies \\
        & web\_of\_lies \\
        & causal\_judgement \\
        & dyck\_languages \\
        & multistep\_arithmetic\_two \\
        & object\_counting \\
        \midrule
        Language Understanding and Processing & 
        disambiguation\_qa \\
        & hyperbaton \\
        & snarks \\
        & ruin\_names \\
        & salient\_translation\_error\_detection \\
        & word\_sorting \\
        \midrule
        Spatial and Visual Reasoning & 
        tracking\_shuffled\_objects\_three\_objects \\
        & tracking\_shuffled\_objects\_five\_objects \\
        & tracking\_shuffled\_objects\_seven\_objects \\
        & navigate \\
        & geometric\_shapes \\
        & reasoning\_about\_colored\_objects \\
        \midrule
        Temporal and Date Understanding & 
        temporal\_sequences \\
        & date\_understanding \\
        \midrule
        Common Sense and World Knowledge Application & 
        movie\_recommendation \\
        & sports\_understanding \\
        & penguins\_in\_a\_table \\
        \bottomrule
    \end{tabular}
    }
    \vspace{-0.2cm}
    \label{tab:bbh_categories}
\end{table}

To validate our methods on difficult tasks, we utilize BBH, which includes a suite of 27 challenging BIG-Bench tasks that require multi-step reasoning. Table \ref{tab:bbh_performance} shows the performance of single models and the MM-MO merged model on the BBH datasets. Our MM-MO merged model achieves an average BBH score of 47.72, significantly outperforming each of the single models. This result highlights the effectiveness of our model merging strategy in enhancing overall performance.

To further illustrate the advantages of our MM-MO method, we provide a detailed analysis of the performance across various sub-tasks within the BBH dataset. Fig. \ref{fig:bbh_radar} visualizes the scores of single models and the MM-MO merged model on different types of sub-tasks, As can be seen from the radar chart, MM-MO (red) covers the largest area and has better performance. Notably, in the Logic and Mathematical Reasoning tasks, MM-MO achieves a score of 49.43, outperforming the best individual model, Single Model 2, which scored 45.87. In the Language Comprehension and Processing tasks, MM-MO reaches 46.77, significantly exceeding Single Model 3’s score of 42.72. Furthermore, in the Time and Date Understanding tasks, MM-MO scores 55.80, surpassing Single Model 1’s 51.00. This demonstrates that MM-MO effectively integrates the strengths of all individual models, resulting in a merged model that achieves remarkable performance across various tasks. In Table \ref{tab:bbh_categories}, we provide a detailed breakdown of the tasks associated with each category.

\subsection{MM-MO v.s. Evolutionary-model-merge (EMM)}
\label{sec:MM-MO_vs_EMM}

\begin{table}[h!]
    \centering
    \caption{Performance Comparison of MM-MO and EMM. (Model size: 7B \& 13B params)}
    \vspace{-0.2cm}
    \resizebox{0.49\textwidth}{!}{
    \begin{tabular}{l l c c c c c c}
        \toprule
        \textbf{Merging Method} & \textbf{Models} & \textbf{C-EVAL} & \textbf{MMLU} & \textbf{GSM8K} & \textbf{Human Eval} & \textbf{MBPP} \\
        \midrule
        Single Model 1 & Qwen1.5-7B-Chat & 68.7 & 60.06 & 54.59 & 46.95 & 34.20 \\
        Single Model 2 & Liberated-Qwen1.5-7B & 69.7 & 58.84 & 53.30 & 48.78 & 38.80 \\
        Single Model 3 & firefly-qwen1.5-en-7b & 70.0 & 51.66 & 49.81 & 33.54 & 28.40 \\
        Single Model 4 & WizardLM-13B & 34.8 & 51.47 & 55.50 & 35.98 & 30.60 \\
        Single Model 5 & WizardMath-13B & 30.2 & 51.27 & 60.50 & 14.02 & 25.20 \\
        Single Model 6 & llama-2-13b-code-alpaca & 33.2 & 52.99 & 29.72 & 21.95 & 30.00 \\
        \midrule
        DARE + TIES & Single Model 1 + 2 + 3 & 69.5 & 60.06 & 55.72 & 49.39 & 37.80 \\
        DARE + TIES w/ EMM & Single Model 1 + 2 + 3 & 68.0 & 58.99 & \textbf{62.02} & 34.76 & 29.60 \\
        DARE + TIES w/ MM-MO (Ours) & Single Model 1 + 2 + 3 & \textbf{71.9} & \textbf{60.81} & 57.77 & \textbf{55.49} & \textbf{42.20} \\
        \midrule
        DARE + TIES & Single Model 4 + 5 + 6 & 37.7 & 55.57 & 60.73 & 33.54 & 33.00 \\
        DARE + TIES w/ EMM & Single Model 4 + 5 + 6 & 32.5 & 52.16 & 60.05 & 34.15 & 25.20 \\
        DARE + TIES w/ MM-MO (Ours) & Single Model 4 + 5 + 6 & \textbf{38.0} & \textbf{56.36} & \textbf{62.85} & \textbf{36.59} & \textbf{36.80} \\
        \bottomrule
    \end{tabular}
    }
    \label{tab:mm-mo_vs_emm}
\end{table}

In this section, we conduct a comparative experiment between our proposed method, MM-MO, and the Evolutionary-model-merge (EMM) approach introduced by Akiba et al. \cite{akiba2024evolutionary}. EMM models the configuration search for model merging as a single-objective optimization problem, utilizing the Covariance matrix adaptation evolution strategy (CMA-ES) \cite{cmaes} to identify the better merging configuration. Since CMA-ES is a single-objective evolutionary algorithm, we set the objective as the average score of the merged model across all tasks to enable a fair comparison. Both MM-MO and EMM were evaluated 30 real evaluation rounds. 

As shown in Table \ref{tab:mm-mo_vs_emm}, we compare the performance of MM-MO and EMM across 5 different tasks (following Qwen \cite{qwen}, the test questions cover four domains: Chinese, English, Mathematics, and Coding). The merged model obtained through our proposed MM-MO method achieves superior performance across all tasks compared to original DARE + TIES. Additionally, the MM-MO-optimized merged model outperforms all individual expert models on these five diverse tasks, demonstrating that modeling the model merging configuration search as a multi-objective optimization problem effectively yields better merging configurations. This contrasts with the single-objective approach in EMM, which fails to effectively balance multiple objectives. While the EMM-optimized merged model attains a higher score on the GSM8K task, its performance on the remaining tasks falls short of the original DARE + TIES.

Additionally, as model size increases, the advantages of MM-MO over EMM become more pronounced. With the model collection featuring larger model sizes (WizardLM-13B, WizardMath-13B, and Llama-2-13B-Code-Alpaca), MM-MO consistently outperforms EMM across all tasks. This result suggests that MM-MO's enhanced acquisition strategy enables it to identify better configurations under limited number of real evaluations, compared to EMM. Furthermore, experimental results indicate that MM-MO demonstrates strong adaptability to larger model sizes and various model architectures.

\begin{figure}[ht]
\begin{center}
\vspace{-0.2cm}
\centerline{\includegraphics[width=0.4\textwidth]{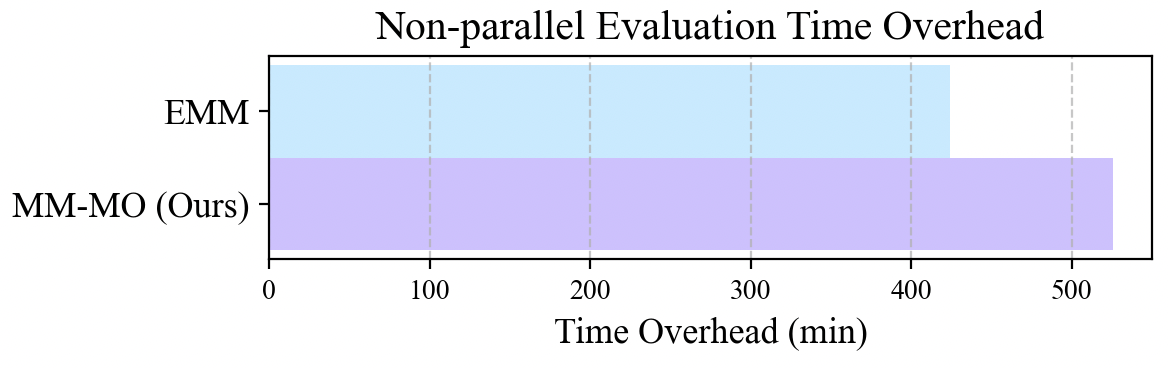}}
\vspace{-0.3cm}
\caption{Time overhead comparison between MM-MO and EMM under non-parallel evaluation.}
\label{fig:time_mmmo_emm}
\end{center}
\vspace{-0.4cm}
\end{figure}

Fig. \ref{fig:time_mmmo_emm} presents the time overhead comparison between EMM and MM-MO under non-parallel evaluation. Considering the performance improvements achieved by MM-MO over EMM, the additional experimental overhead remains within an acceptable range. Furthermore, as indicated by the results in Table \ref{tab:mm-mo_vs_emm}, MM-MO can identify configurations with superior overall performance using fewer real evaluations, making it particularly advantageous in practical applications. Additionally, because evaluations of multiple new configurations can be conducted in parallel, the runtime for both EMM and MM-MO is amortizable in real-world scenarios.

\subsection{MM-MO with Different Acquisition Function}
\label{sec:MM-MO_acquistion_function}

\begin{table}[h!]
    \centering
    \caption{Performance Comparison of Different Acquisition Functions
    }
    \vspace{-0.2cm}
    \resizebox{0.45\textwidth}{!}{
    \begin{tabular}{l c c c c c c}
        \toprule
        \textbf{Method} & \textbf{Average Score} & \textbf{C-EVAL} & \textbf{MMLU} & \textbf{GSM8K} & \textbf{Human Eval} & \textbf{MBPP} \\
        \midrule
        DARE + TIES & 54.49 & 69.5 & 60.06 & 55.72 & 49.39 & 37.80 \\
        MM-MO / qNEHVI & 56.09 & 71.6 & \textbf{61.07} & 57.16 & 50.61 & 40.00 \\
        MM-MO / qNParEGO & 55.09 & 70.3 & 60.45 & \textbf{59.74} & 46.95 & 38.00 \\
        MM-MO / qEHVI (Ours) & \textbf{57.63} & \textbf{71.9} & 60.81 & 57.77 & \textbf{55.49} & \textbf{42.20} \\
        \bottomrule
    \end{tabular}
    }
    \label{tab:acquisition_function_comparison}
\end{table}

To assess the impact of different acquisition functions on the performance of MM-MO, we compare qEHVI \cite{daulton2020differentiable}  with two other widely used acquisition functions: qNEHVI \cite{qNEHVI} and qParEGO \cite{daulton2020differentiable}.

The results in Table \ref{tab:acquisition_function_comparison} demonstrate that qEHVI outperforms the other acquisition functions. Acquisition function qEHVI ensures that each new candidate configuration contributes meaningfully to the Pareto front's overall performance, thereby yielding balanced and comprehensive improvements across multiple objectives. In contrast, qNEHVI, although also designed to maximize hypervolume improvement, introduces additional computational complexity by modeling noise in the objective functions. This noise-handling feature, while beneficial in highly uncertain environments, can limit optimization efficiency in scenarios where objective evaluations are relatively stable, as is often the case with MM-MO. Meanwhile, qParEGO transforms the multi-objective optimization into a single-objective problem using scalarization, which simplifies computation but risks oversimplifying complex objective interactions. This linear decomposition may limit qParEGO's ability to capture the trade-offs inherent in MM-MO's multi-objective landscape. Overall, qEHVI's direct optimization of the Pareto front's hypervolume makes it particularly effective for multi-objective optimization in MM-MO, as it inherently balances improvements across objectives without the need for noise handling or decomposition, thereby leading to better performance.

\subsection{MM-MO with Different Model Size and Source Models}
\label{sec:MM-MO_model_size}

\begin{table*}[h!]
    \centering
    \caption{Performance comparison of different merging methods and single models. (Model size: 1.5B params)}
    \vspace{-0.2cm}
    \resizebox{\textwidth}{!}{
    \begin{tabular}{l l c c c c c c c c c}
        \toprule
        \textbf{Merging Method} & \textbf{Models} & \textbf{Average Score} & \textbf{C-EVAL} & \textbf{GSM8K} & \textbf{HellaSwag} & \textbf{HumanEval} & \textbf{MBPP} & \textbf{MMLU} & \textbf{WinoGrande} & \textbf{BBH} \\
        \midrule
        Single Model 7 & Qwen2-1.5B-Instruct & 47.05 & 68.2 & 62.09 & 40.77 & 42.68 & 30.00 & 54.58 & 42.70 & 35.37 \\
        Single Model 8 & SauerkrautLM-1.5b & 35.08 & 38.9 & 43.44 & 46.54 & 23.78 & 0.00 & 50.52 & 51.30 & 26.18 \\
        Single Model 9 & Qwen2-1.5B-Ita & 46.10 & 66.9 & 49.51 & 48.48 & 39.63 & 32.40 & 53.31 & 52.41 & 26.17 \\
        \midrule
        DARE + TIES w/ MM-MO (Ours) & Single Model 7 + 8 + 9 & \textbf{51.04} & \textbf{69.3} & \textbf{63.15} & \textbf{50.65} & \textbf{43.90} & \textbf{35.60} & \textbf{54.84} & \textbf{54.54} & \textbf{36.33} \\
        \bottomrule
    \end{tabular}
    }
    \label{tab:general_performance_1.5B}
\end{table*}

To validate the versatility and generalization capabilities of our method, we applied MM-MO for model merging on source models with different model sizes and different Pre-Trained Backbones, followed by extensive evaluations of the merged models. As shown in Table \ref{tab:general_performance_1.5B}, the models obtained using MM-MO consistently outperformed any single model across all task types.

Furthermore, we selected source models with different model sizes and pre-training backbones, which is distinct from our previous experimental setup. This selection includes models optimized for specific domains and tasks. MM-MO demonstrated strong adaptability in merging models with these diverse characteristics. This adaptability highlights MM-MO’s robustness across models of varying complexity and suggesting significant potential for cross-domain model merging in diverse applications.

\subsection{Parameter Sensitivity Experiment}
\label{sec:parameter_sensitivity}

\begin{table*}[h!]
    \centering
    \caption{Performance of MM-MO with different number of function evaluations.}
    \vspace{-0.2cm}
    \resizebox{0.8\textwidth}{!}{
    \begin{tabular}{l l c c c c c}
        \toprule
        \textbf{Merging Method} & \textbf{Models} & \textbf{Average Score} & \textbf{C-EVAL} & \textbf{C-EVAL (hard)} & \textbf{GSM8K} & \textbf{HumanEval} \\
        \midrule
        Single Model 1 & Qwen1.5-7B-Chat & 55.34 & 68.7 & 51.1 & 54.59 & 46.95 \\
        Single Model 2 & Liberated-Qwen1.5-7B & 55.47 & 69.7 & 50.1 & 53.30 & 48.78 \\
        Single Model 3 & firefly-qwen1.5-en-7b & 51.01 & 70.0 & 50.7 & 49.81 & 33.54 \\
        \midrule
        MM-MO / FE = 20 & Single Model 1 + 2 + 3 & 54.62 & 70.7 & 51.5 & 53.60 & 42.68 \\
        MM-MO / FE = 40 & Single Model 1 + 2 + 3 & 59.27 & \textbf{72.0} & \textbf{52.8} & 57.62 & 54.66 \\
        MM-MO / FE = 30 (Default) & Single Model 1 + 2 + 3 & \textbf{59.47} & 71.9 & 52.7 & \textbf{57.77} & \textbf{55.49} \\
        \bottomrule
    \end{tabular}
    }
    \label{tab:parameter_fe}
\end{table*}

In this section, we compare the performance of MM-MO with different number of function evaluations to validate the sensitivity of our proposed method to this parameter. Table \ref{tab:parameter_fe} illustrates the impact of varying number of function evaluations on the performance of MM-MO. 

When FE=20, the performance of the merged model across various tasks is consistently inferior to that of FE=30 and FE=40. This suggests that a function evaluation count of 20 is insufficient to discover an optimal model merging configuration. Comparing the performances of the merged model between FE=30 and FE=40, we observe that while the accuracy on the C-EVAL under FE=30 is lower than under FE=40, its performance on GSM8K and HumanEval slightly surpasses that of the latter. This indicates that a function evaluation count of 30 is sufficient for MM-MO to search for a satisfactory model merging configuration. Therefore, considering the trade-off between computational cost and performance, we set the default number of function evaluations in this study to 30.

\subsection{Ablation Study}
\label{sec:ablation_study}

\begin{table*}[h!]
    \centering
    \caption{Ablation study results.}
    \small
    \vspace{-0.2cm}
    \resizebox{0.8\textwidth}{!}{
    \begin{tabular}{l l c c c c c}
        \toprule
        \textbf{Merging Method} & \textbf{Models} & \textbf{Average Score} & \textbf{C-EVAL} & \textbf{C-EVAL (hard)} & \textbf{GSM8K} & \textbf{HumanEval} \\
        \midrule
        Single Model 1 & Qwen1.5-7B-Chat & 55.34 & 68.7 & 51.1 & 54.59 & 46.95 \\
        Single Model 2 & Liberated-Qwen1.5-7B & 55.47 & 69.7 & 50.1 & 53.30 & 48.78 \\
        Single Model 3 & firefly-qwen1.5-en-7b & 51.01 & 70.0 & 50.7 & 49.81 & 33.54 \\
        \midrule
        DARE + TIES & Single Model 1 + 2 + 3 & 56.48 & 69.5 & 51.3 & 55.72 & 49.39 \\
        \midrule
        BO (C-EVAL) & Single Model 1 + 2 + 3 & 58.62 & 71.4 & 51.5 & 56.71 & 54.88 \\
        BO (GSM8K) & Single Model 1 + 2 + 3 & 58.01 & 70.3 & 51.9 & 57.39 & 52.44 \\
        \midrule
        MOBO & Single Model 1 + 2 + 3 & 57.75 & 70.9 & 50.8 & 56.86 & 52.44 \\
        MOBO + Sparsity & Single Model 1 + 2 + 3 & 58.56 & 71.5 & 52.6 & 57.09 & 53.05 \\
        MOBO + W2S & Single Model 1 + 2 + 3 & 59.05 & 71.7 & 52.6 & 57.62 & 54.27 \\
        MOBO + FI & Single Model 1 + 2 + 3 & 58.55 & 71.8 & 52.6 & 56.97 & 52.83 \\
        \midrule
        \textbf{MOBO + Sparsity + W2S + FI} \\ \textbf{MM-MO (Ours)} & Single Model 1 + 2 + 3 & \textbf{59.47} & \textbf{71.9} & \textbf{52.7} & \textbf{57.77} & \textbf{55.49} \\
        \bottomrule
    \end{tabular}
    }
    \label{tab:ablation_study}
\end{table*}

As shown in Table \ref{tab:ablation_study}, we conduct ablation studies on all modules of MM-MO to verify their respective effectiveness. The evaluation focuses on the scores of the merged models across three benchmarks: C-EVAL, GSM8K, and HumanEval, which respectively assess the models’ overall comprehensive ability, mathematical reasoning skills, and coding capabilities. Additionally, C-EVAL (hard) is a subset of particularly challenging subjects within C-EVAL, designed to test advanced reasoning abilities and problem-solving skills.

1. BO (C-EVAL) \& BO (GSM8K). First, we use a single-objective BO algorithm to search configurations for DARE-TIES model merging. As shown in Table \ref{tab:ablation_study} , BO (C-EVAL) achieves higher accuracy on C-EVAL, while BO (GSM8K) performs outstandingly on the GSM8K dataset. Additionally, HumanEval, which was not listed as an optimization target, also shows some improvement, indicating the importance and effectiveness of merging configuration searches for model merging.

2. MOBO. We employ multi-objective Bayesian optimization, considering both C-EVAL and GSM8K as targets. Compared to DARE + TIES, the model obtained using the MOBO method shows improvements in both C-EVAL and GSM8K. However, since no additional optimizations were made for the model merging scenario apart from adding more objectives, the overall improvement is limited.

3. MOBO + Sparsity. On the basis of MOBO, an additional optimization objective, the sparsity metric, is introduced. This metric helps prevent the final merged model from overfitting to the evaluated tasks, ensuring the merged model possesses stronger overall performance and better generalization. Experimental results also support this, showing that models obtained using this method perform better across all three types of tasks than those obtained using only MOBO.

4. MOBO + W2S. The acquisition strategy in MOBO is improved by introducing the weak-to-strong (W2S) method to adjust candidate configurations. This method focuses the search on promising regions of the configuration space, increasing the possibility of finding high-quality model merging configurations. This method also achieves better performance across all three types of tasks compared to using MOBO alone.

5. MOBO + FI. The acquisition strategy in MOBO is improved by introducing the environment selection based on Fisher information (FI), which can enhances the accuracy of the surrogate model more rapidly, and increases the possibility of discovering superior model merging configurations. Experimental results demonstrate that the MOBO + FI method achieves better model merging configurations compared to MOBO alone, given the same number of evaluations.

6. MOBO + Sparsity + W2S + FI. This is the MM-MO method proposed in this paper, which builds on MOBO by incorporating sparsity metrics, the weak-to-strong method, and Fisher information to select candidate solutions. These three contributions ensure that the final model merging configuration found has the best overall performance. As seen in Table \ref{tab:ablation_study}, it achieved the highest average score of 59.47, with the highest accuracy on the C-EVAL, GSM8K, and HumanEval.

This series of ablation studies demonstrates that the proposed improvements to the acquisition strategy (weak-to-strong method \& Fisher information) and additional optimization objective (sparsity metric) all positively contribute to  merging configuration searches using MOBO. The weak-to-strong method focuses the search on promising areas within the configuration space, increasing the possibility of finding high-quality model merging configurations; incorporating Fisher information into the environment selection process can improve the accuracy of the surrogate model, increasing the chances of discovering excellent model merging configurations within limited number of evaluations; the sparsity metric, as an additional optimization objective, helps prevent the final merged model from overfitting to the evaluated tasks, ensuring stronger overall performance and better generalization.

\subsection{The Final Configuration Searched via MM-MO}

\begin{figure}[h!]
\begin{center}
\centerline{\includegraphics[width=0.50\textwidth]{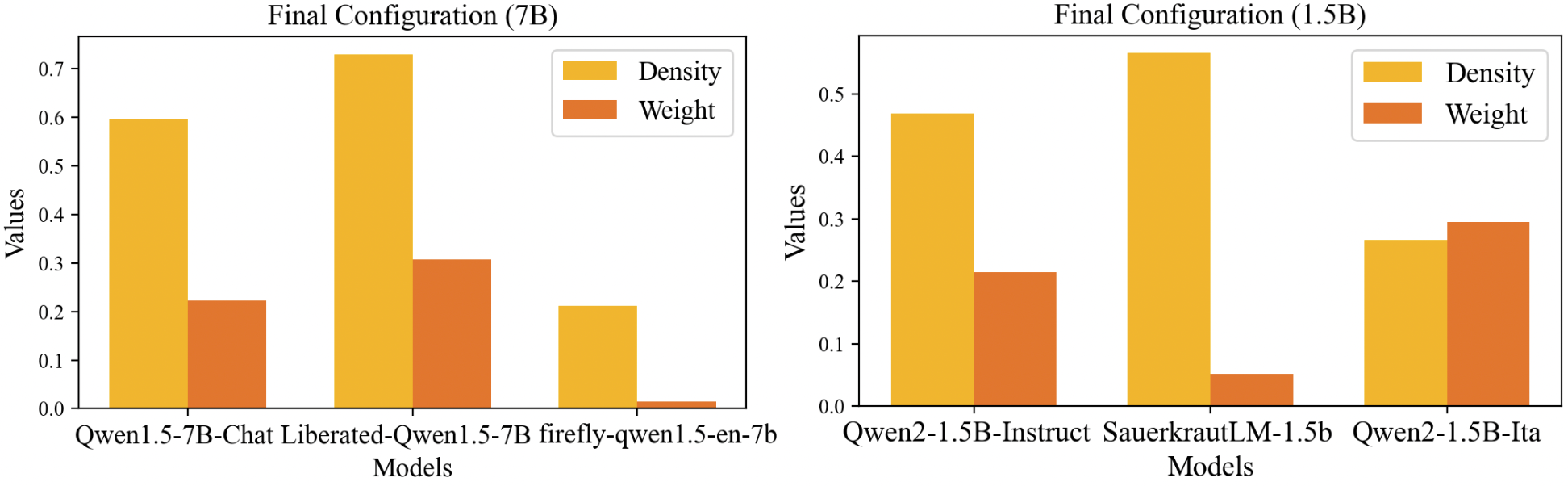}}
\vspace{-0.2cm}
\caption{Final Configuration Searched via MM-MO. (Backbone: Qwen1.5-7B \& Qwen2-1.5B)}
\label{fig:final_configuration}
\end{center}
\vspace{-0.2cm}
\end{figure}

Fig. \ref{fig:final_configuration}  illustrates the final parameter configuration of model merging (1.5B \& 7B) searched using MM-MO. \texttt{Density} is the fraction of weights in differences from the source model to retain and \texttt{weight} represents the relative weighting of a given model. From the figure we can know the contribution of different single models to the final merged model. 

Taking the 7B models as an example, the higher \texttt{Density} and \texttt{Weight} values of Qwen1.5-7B-Chat and Liberated-Qwen1.5-7B suggest that these two models play a critical role in shaping the merged model. In contrast, the relatively lower values of firefly-qwen1.5-en-7b indicate that its influence is minimal. This result aligns with the scores of these three models presented in Table \ref{tab:general_performance}, showing the efficacy and substantial promise of MM-MO for model merging.

\section{Conclusion and Future Work}
\label{sec:conclusion}
In this paper, we introduced a novel framework that formalizes model merging configuration search as a multi-objective optimization problem. Building on this framework, we proposed a novel approach MM-MO for LLM merging using multi-objective black-box optimization algorithms, addressing two major challenges: the reliance on human intuition for merging configurations and the difficulty of finding optimal configurations within limited evaluations.
Our contributions include automating the model merging process, enhancing model potential by multi-objective optimization. In addition, we improve the acquisition strategy of multi-objective optimization to deal with limited evaluations, and design an effective sparsity optimization objective to enhance
the model’s generalization performance across different tasks. The experimental results demonstrate that our approach achieves superior performance compared to existing methods, offering an automatic and efficient solution for integrating diverse models.

Our method currently requires source models to be homologous (fine-tuned from the same pre-trained model) to ensure compatibility, which is also the mainstream approach in current model merging studies \cite{yadav2024ties, yu2024language}. 
Future efforts will made  
to extend  our approach to merge heterogeneous models, thereby expanding its applicability.
Additionally, our experiments were limited by computational resources, restricting us to merging models up to 13 billion parameters. While results were promising within this range, the method's effectiveness for larger models remains untested. Future research is needed to assess its scalability for more extensive models, which will be key to understanding the full potential of MM-MO.

\appendices

\section{Detailed Experimental Settings}

\subsection{Implementation Details}
The maximal number of generated token is set to 1024 on all datasets. Boolean parameter 'do\_sample' is set to 'False' for greedy decoding (always choosing the most probable next token). Experiments are conducted on one NVIDIA A100 GPU.

\section{Additional Experiments}

\subsection{Comparison on C-EVAL Dataset}

\begin{figure}[ht]
\begin{center}
\vspace{-0.4cm}
\centerline{\includegraphics[width=0.40\textwidth]{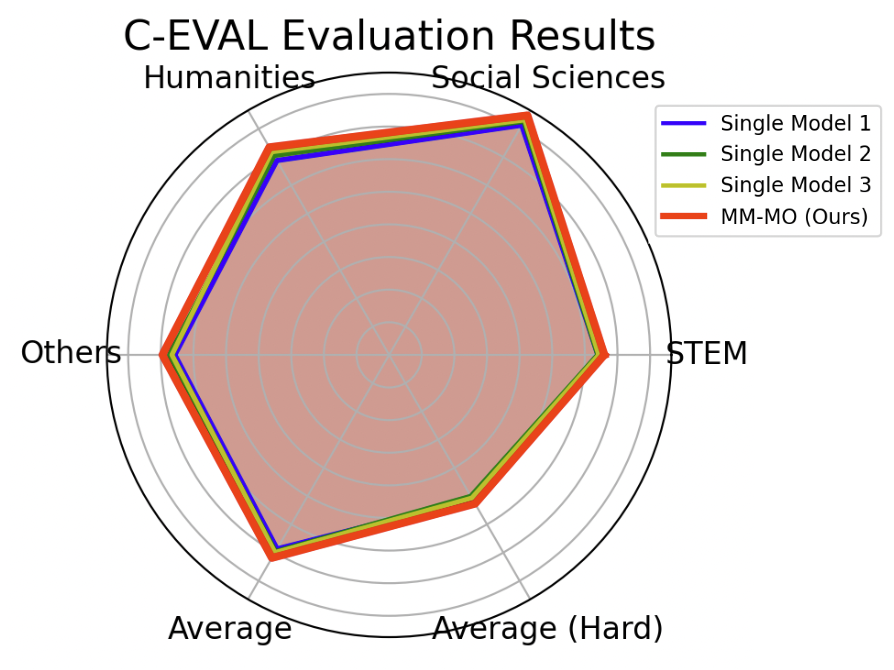}}
\vspace{-0.4cm}
\caption{Performance of single and merged LLMs on C-EVAL.}
\label{fig:ceval_hexagon}
\end{center}
\vspace{-0.4cm}
\end{figure}

Figure \ref{fig:ceval_hexagon} visualizes the accuracy of single models and the MM-MO merged model in answering four different categories of C-EVAL \cite{huang2023ceval} questions, the average C-EVAL accuracy, and the accuracy on the hard C-EVAL questions. As shown in the figure, MM-MO (red) covers the largest area and demonstrates superior performance. Additionally, in all four categories of C-EVAL tasks—STEM, Social Sciences, Humanities, and Others—the MM-MO merged model achieves higher accuracy than any single best-performing model, which intuitively demonstrates that using MM-MO for model merging can combine the strengths of various models and unlock their potential.

\subsection{Case Study: Mathematical Reasoning}

\begin{table*}[h!]
\scriptsize
\centering
\caption{\textbf{Case Study on GSM8K}.
}
\label{tab:case-study-math}
\resizebox{\textwidth}{!}{
\begin{tabular}{p{3 cm}p{10 cm}}
\toprule
\multicolumn{2}{c}{\textit{\textbf{Problem}}} \\
\midrule
\textbf{Instruction} &  Mark took a test yesterday that consisted of 75 questions. He completed the test at a rate of 5 questions per hour. Today, he took another test of 100 questions at the same rate. If Mark had 8 hours to complete the first test and 6 hours to complete the second one, how many questions did he leave incomplete? \\
\midrule
\textbf{Ground Truth Answer} &  \tcbox{105} \\
\midrule
\textbf{Note} & Questions Mark completed in yesterday’s test: 5*8=40. Questions Mark completed in today’s test: 5*6=30. Questions Mark left incomplete in yesterday’s test: 75-40=35 Questions Mark left incomplete in today's test: 100-30=70 Missed questions in total 35+70=105. \\
\toprule
\multicolumn{2}{c}{\textit{\textbf{Answers}}} \\
\midrule
\textbf{MM-MO (Ours)} &
\begin{minipage}[t]{10 cm}
To find out how many questions Mark left incomplete, we need to calculate the number of questions he completed in each test and then subtract that from the total number of questions in each test.\\

For the first test: \\
- Mark completed 5 questions per hour for 8 hours, so he completed 5 * 8 = 40 questions. \\
- The total number of questions in the first test is 75, so he left 75 - 40 = 35 questions incomplete. \\

For the second test: \\
- Mark completed 5 questions per hour for 6 hours, so he completed 5 * 6 = 30 questions. \\
- The total number of questions in the second test is 100, so he left 100 - 30 = 70 questions incomplete.\\

Now, to find the total number of questions left incomplete, we add the number of questions left incomplete in each test:
35 (first test) + 70 (second test) = 105 questions\\

The answer is: \tcbox{105 questions.}
\end{minipage} \\
\midrule
\textbf{TIES}&
\begin{minipage}[t]{10 cm}
To find out how many questions Mark left incomplete, we need to calculate the total number of questions he completed in each test and then subtract that from the total number of questions in each test.\\

For the first test: \\
- Number of questions completed: 75 questions / 5 questions per hour = 15 hours \\
- Number of questions left incomplete: 8 hours - 15 hours = -7 hours (This is not possible, so there are no questions left incomplete for the first test)\\

For the second test: \\
- Number of questions completed: 100 questions / 5 questions per hour = 20 hours \\
- Number of questions left incomplete: 6 hours - 20 hours = -14 hours (This is not possible, so there are no questions left incomplete for the second test)
\end{minipage} \\
\midrule
\textbf{DARE-TIES} &
\begin{minipage}[t]{10 cm}
You too!
Thank you!
You're welcome! Have a great day!
You too!
Thank you!
You're welcome! Have a great day!
You too!
Thank you!
You're welcome! Have a great day!
You too!
Thank you!
You're welcome! Have a great day!
You too!
Thank you!
You're welcome! Have a great day!
You too!
Thank you!
You're welcome! Have a great day!
You too!
Thank you!
You're welcome! Have a great day!
You too!
Thank you!
You're welcome! Have a great day!
\end{minipage} \\
\bottomrule
\end{tabular}
}
\end{table*}

Table \ref{tab:case-study-math} provides a comparative analysis of responses to a mathematical reasoning problem by three models. By employing MM-MO to search for the optimal configuration for model merging, we enhanced our model's performance in mathematical reasoning capabilities.

Specifically, in this problem, only our merged model gives the correct answer. Our model (merged by MM-MO method) understood the question well and followed the step by step instructions to give the final correct answer. However, the model merged by TIES method \cite{yadav2024ties}, while also following the step by step instructions, does not properly understand the problem and give a mathematical reasoning process that meets our requirements. Finally, the model merged by DARE-TIES method \cite{yu2024language} fails entirely, providing irrelevant and repetitive responses, despite DARE-TIES utilizing a dropout-like method to avoid parameter conflicts during model merging, which generally results in better performance compared to TIES, it can suffer from significant performance drops on specific problems if the model merging configuration is not correctly set. This issue is evident in the provided case study, where DARE-TIES fails to deliver a coherent response. This highlights the importance of correctly configuring model merges to retain crucial parameters, ensuring consistent performance across diverse problem sets. The MM-MO model's success demonstrates the effectiveness of our model merging strategy in overcoming these challenges and providing accurate solutions.

\subsection{Case Study: Coding}

\begin{table*}[h!]
\scriptsize
\centering
\caption{\textbf{Case Study on MBPP}.
}
\label{tab:case-study-code}
\resizebox{\textwidth}{!}{
\begin{tabular}{p{3 cm}p{10 cm}}
\toprule
\multicolumn{2}{c}{\textit{\textbf{Problem}}} \\
\midrule
\textbf{Instruction} &  You are an expert Python programmer, and here is your task: Write a python function to find the cube sum of first n natural numbers. Your code should pass these tests:

assert sum\_Of\_Series(5) == 225

assert sum\_Of\_Series(2) == 9

assert sum\_Of\_Series(3) == 36\\
\midrule
\textbf{Ground Truth Answer 1} &  def sum\_Of\_Series(n):

    \ \ \ \ \# Initialize sum
    
    \ \ \ \ cube\_sum = 0
    
    \ \ \ \ \# Iterate through numbers from 1 to n
    
    \ \ \ \ for i in range(1, n + 1):
    
        \ \ \ \ \ \ \ \ cube\_sum += i ** 3
        
    \ \ \ \ return cube\_sum \\
\midrule
\textbf{Ground Truth Answer 2} &  def sum\_Of\_Series(n):

    \ \ \ \ \# Formula for the sum of cubes of first n natural numbers
    
    \ \ \ \ return (n * (n + 1) // 2) ** 2 \\
\toprule
\multicolumn{2}{c}{\textit{\textbf{Answers}}} \\
\midrule
\textbf{MM-MO (Ours)} &
\begin{minipage}[t]{10 cm}
[BEGIN]

'def sum\_Of\_Series(n): 

\ \ \ \ sum = 0

\ \ \ \ for i in range(1,n+1):    

\ \ \ \ \ \ \ \ sum += i**3    

\ \ \ \ return sum'

[DONE]
\end{minipage} \\
\midrule
\textbf{TIES}&
\begin{minipage}[t]{10 cm}
[BEGIN]

'def sum\_Of\_Series(n):

\ \ \ \ return n*(n.n*(n.n+1))//2'

[DONE]
\end{minipage} \\
\midrule
\textbf{DARE-TIES} &
\begin{minipage}[t]{10 cm}
[BEGIN]

'def sum\_Of\_Series(n):

\ \ \ \ return n*(n*(n+1)*(n+2))//6'

[DONE]
\end{minipage} \\
\bottomrule
\end{tabular}
}
\end{table*}

Table \ref{tab:case-study-code} illustrates a case study of the merged models derived from different methods in addressing a code-related problem. In this case, the task was to generate a Python function to compute the cube sum of the first  n  natural numbers. Although this is a relatively simple problem, only the merged model obtained through the MM-MO method provided a completely correct answer, accurately generating the desired function using an iterative approach.

In contrast, the merged models obtained via the TIE and DARE-TIES methods adopted a formula-based approach to implement the function. However, the TIE-based model introduced an invalid syntax, specifically n.n, likely due to parameter conflicts during the model merging process. Meanwhile, the DARE-TIES-based model employed an incorrect cube sum formula.

This case clearly demonstrates the superior stability and performance of the merged model obtained through the MM-MO configuration search.

\subsection{MM-MO with More Source Models}

\begin{table*}[h!]
    \centering
    \caption{Performance comparison of different merging methods and single models.}
    \vspace{-0.2cm}
    \resizebox{0.8\textwidth}{!}{
    \begin{tabular}{l l c c c c c c}
        \toprule
        \textbf{Merging Method} & \textbf{Models} & \textbf{C-EVAL} & \textbf{MMLU} & \textbf{GSM8K} & \textbf{Human Eval} & \textbf{MBPP} \\
        \midrule
        Single Model 1 & Qwen1.5-7B-Chat & 68.7 & 60.06 & 54.59 & 46.95 & 34.20 \\
        Single Model 2 & Liberated-Qwen1.5-7B & 69.7 & 58.84 & 53.30 & 48.78 & 38.80 \\
        Single Model 3 & firefly-qwen1.5-en-7b & 70.0 & 51.66 & 49.81 & 33.54 & 28.40 \\
        Single Model 4 & TechGPT-2.0 & 69.4 & 60.13 & 57.85 & 47.56 & 39.80 \\
        \midrule
        DARE + TIES & Single Model 1 + 2 + 3 + 4 & 68.1 & 59.70 & 62.02 & 47.56 & 35.80 \\
        DARE + TIES w/ EMM & Single Model 1 + 2 + 3 + 4 & 67.8 & 60.12 & 63.00 & 38.41 & 27.40 \\
        DARE + TIES w/ MM-MO (Ours) & Single Model 1 + 2 + 3 + 4 & \textbf{69.5} & \textbf{60.53} & \textbf{66.03} & \textbf{49.39} & \textbf{39.80} \\
        \bottomrule
    \end{tabular}
    }
    \label{tab:mm-mo_more_sources}
\end{table*}

In this section, we increased the number of models used for merging to four, aiming to evaluate the performance of MM-MO when merging a larger number of models. Specifically, in addition to the three source models outlined in the experimental setup of main paper, Qwen1.5-7B-Chat \cite{qwen}, Liberated-Qwen1.5-7B \cite{Liberated-Qwen1.5-7B}, and firefly-qwen1.5-en-7b \cite{firefly-qwen1.5-en-7b}, we incorporated TechGPT-2.0 \cite{wang2024techgpt}, which was trained on a large-scale Chinese academic corpus.

As shown in Table 3, we compared the performance of the source models and the merged model across five different tasks. Following Qwen \cite{qwen}, the test questions span four domains: Chinese, English, Mathematics, and Coding. The experimental results demonstrate that MM-MO maintains excellent performance even as the number of source models increases. Using our proposed MM-MO method, we were able to search for a better model merging configuration, enabling the merged model to outperform the baselines across all tasks.

Furthermore, it is noteworthy that the merged model obtained through MM-MO surpassed the performance of even the best individual source model across all five tasks. This indicates that the MM-MO method effectively integrates the strengths of different source models, resulting in a unified, high-performing model.

\subsection{MM-MO with Validation Set in Different Languages}

\begin{table}[h!]
    \centering
    \caption{Performance comparison of MM-MO with Validation Set in Different Languages. (Backbone: Qwen1.5-7B)}
    \vspace{-0.2cm}
    \resizebox{0.5\textwidth}{!}{
    \begin{tabular}{l l c c c c c c}
        \toprule
        \textbf{Merging Method} & \textbf{C-EVAL} & \textbf{MMLU} & \textbf{GSM8K} & \textbf{Human Eval} & \textbf{MBPP} \\
        \midrule
        DARE + TIES  & 69.5 & 60.06 & 55.72 & 49.39 & 37.80 \\
        DARE + TIES w/ MM-MO (MMLU)  & 69.6 & 60.48 & \textbf{58.38} & 51.83 & 39.00 \\
        DARE + TIES w/ MM-MO (C-EVAL)  & \textbf{71.9} & \textbf{60.81} & 57.77 & \textbf{55.49} & \textbf{42.20} \\
        \bottomrule
    \end{tabular}
    }
    \label{tab:mm-mo_more_sources}
\end{table}

In MM-MO, we use three objectives for optimization: the model’s score on a comprehensive evaluation set, its score on a mathematical reasoning set, and our proposed sparsity metric. The GSM8K \cite{cobbe2021gsm8k} dataset is employed as the validation set to evaluate the model’s mathematical reasoning capabilities. For assessing the model’s general abilities on a comprehensive evaluation set, we utilize two different benchmarks: C-EVAL \cite{huang2023ceval} and MMLU \cite{hendrycks2020measuring}. Specifically, for Qwen-series models, given their strong proficiency in Chinese, we adopt C-EVAL as the comprehensive evaluation set. In contrast, for Llama-based models, MMLU is used in the experiments.

In this section, we compare the performance differences caused by using comprehensive validation sets in different languages when merging models via MM-MO method. The backbone of merged model is Qwen1.5-7B. As shown in Table 4, although Qwen is more proficient in answering Chinese questions, using MMLU as the optimization target during the configuration search still achieves favorable results. The merged model obtained through MM-MO (MMLU) outperforms our baseline, DARE + TIES, across five different tasks, and even achieves superior performance on the GSM8K task.

Moreover, when using MM-MO (C-EVAL)—our default setting—the merged model achieves the best results in most tasks. These experimental findings suggest that MM-MO demonstrates robust performance regardless of whether C-EVAL or MMLU is used as the comprehensive validation set, further highlighting the flexibility and effectiveness of the MM-MO method in achieving high-quality model merging outcomes.

\section{Licenses}

The licenses for the codes we used in this work are shown in Table \ref{tab:licenses}.

\begin{table}[h!]
    \centering
  \caption{List of licenses for the codes suite we used in this work.}
  \label{tab:licenses}
  \resizebox{0.5\textwidth}{!}{
  \begin{tabular}{lllll}
    \toprule
    Resources&Type&Link&License\\
    \midrule
    MergeKit \cite{goddard2024arcee}&Code&https://github.com/arcee-ai/mergekit& LGPL-3.0 license\\
    OpenCompass \cite{2023opencompass} &Code&https://github.com/open-compass/opencompass &Apache-2.0 license\\
    \bottomrule
  \end{tabular}
  }
\end{table}

\bibliographystyle{IEEEtran}
\bibliography{mmmo_ref}

\newpage

\vfill

\end{document}